# Vision-Language and Large Language Model Performance in Gastroenterology: GPT, Claude, Llama, Phi, Mistral, Gemma, and Quantized Models

**Running title:** LLM and VLM Capabilities on Gastroenterology Board Exams

Seyed Amir Ahmad Safavi-Naini[1,2,3†], Shuhaib Ali[4†], Omer Shahab[5†], Zahra Shahhoseini[6], Thomas Savage [7], Sara Rafiee[6], Jamil S. Samaan[8], Reem Al Shabeeb[9], Farah Ladak[4], Jamie O. Yang[10], Juan Echavarria[4], Sumbal Babar[11], Aasma Shaukat[12], Samuel Margolis[13], Nicholas P. Tatonetti [14,15,16], Girish Nadkarni [1,2], Bara El Kurdi[4*]**,** Ali Soroush[1,2,3*]

1. Division of Data-Driven and Digital Health (D3M), The Charles Bronfman Institute for Personalized Medicine, Icahn School of Medicine at Mount Sinai, New York, NY, USA
2. The Charles Bronfman Institute of Personalized Medicine, Icahn School of Medicine at Mount Sinai, New York, NY, USA;
3. Henry D. Janowitz Division of Gastroenterology, Icahn School of Medicine at Mount Sinai, New York, New York, USA
4. University of Texas Health, Department of Medicine, Division of Gastroenterology and Hepatology, San Antonio, Texas, USA
5. Division of Gastroenterology, Virginia Hospital Center, Arlington, Virginia, USA
6. School of Medicine, Shahid Beheshti University of Medical Sciences, Tehran, Iran
7. Department of Medicine, Stanford University, Stanford, California
8. Karsh Division of Gastroenterology and Hepatology, Cedars-Sinai Medical Center, Los Angeles, CA
9. Department of Medicine, Inova Fairfax Medical Campus, Falls Church, Virginia, USA.
10. University of California–Los Angeles, Department of Internal Medicine, Los Angeles, CA, USA.
11. University of Texas Health, Department of Medicine, Division of Infectious Diseases, San Antonio, Texas, USA.
12. Division of Gastroenterology, Department of Medicine, NYU Grossman School of Medicine, New York, NY, USA.
13. David Geffen School Medicine, University of California, Los Angeles, CA, USA
14. Department of Computational Biomedicine, Cedars-Sinai Medical Center, West Hollywood, California, USA
15. Cedars-Sinai Cancer, Cedars-Sinai Medical Center, 8700 Beverly Blvd. Los Angeles, CA, USA
16. Department of Biomedical Informatics, Columbia University, New York, New York, USA.

† Shuhaib Ali, Omer Shahab, and Seyed Amir Ahmad Safavi-Naini contributed equally to this work.
\* **Correspondence to**: Ali Soroush (Ali.Soroush@mountsinai.org) and Bara El Kurdi (baraabayern@gmail.com),

**Funding:** None
**COI Disclosures:** SAASN: Received nonsignificant financial compensation in 2023 as an R&D associate of Aryasp Co.; SA: None; OS: None; ZS: None; TS: None; SR: None; JSS: None; RAS: None; FL: None; JOY: None; JE: None; SB: None; ASh: None; SM: None; NPT: None; GN: None; BEK: None; ASo: None
Author Contributions:
**Data Transparency Statement:** Code is available at: https://github.com/Sdamirsa/LLM-VLM-in-Gastroenterology . Study data from American College of Gastroenterology (ACG) are restricted and available upon request with ACG permission.
**Manuscript Pages:** 34; **Figures**: 7; **Tables**: 2; **Supplementary File Pages**: 35;

# Author details

| Full Name | Email; ORCID | Degree; Position | Affiliation |
|---|---|---|---|
| Seyed Amir Ahmad Safavi-Naini | sdamirsa@ymail.com; 0000-0001-9295-9283 | MD; Postdoc Fellow | 1- Division of Data-Driven and Digital Health (D3M), The Charles Bronfman Institute for Personalized Medicine, Icahn School of Medicine at Mount Sinai, New York, NY, USA; 2- The Charles Bronfman Institute of Personalized Medicine, Icahn School of Medicine at Mount Sinai, New York, NY, USA; 3- Henry D. Janowitz Division of Gastroenterology, Icahn School of Medicine at Mount Sinai, New York, New York, USA |
| Shuhaib Ali | alis9@uthscsa.edu | MD | University of Texas Health, Department of Medicine, Division of Gastroenterology and Hepatology, San Antonio, Texas, USA |
| Omer Shahab | omershahab@gmail.com | MD | Division of Gastroenterology, Virginia Hospital Center, Arlington, Virginia, USA |
| Zahra Shahhoseini | zshahhoseini2020@gmail.com; 0009-0009-6980-4174 | BS, Medical Student | School of Medicine, Shahid Beheshti University of Medical Sciences, Tehran, Iran |
| Thomas Savage | tsavage@stanford.edu; 0000-0003-4828-5802 | MD | Department of Medicine, Stanford University, Stanford, California |
| Sara Rafiee | Sara.rfe81@gmail.com; 0009-0008-5180-7388 | Medical Student | School of Medicine, Shahid Beheshti University of Medical Sciences, Tehran, Iran |
| Jamil S. Samaan | jamil.samaan@cshs.org; 0000-0002-6191-2631 | MD; Gastroenterology clinical fellow | Karsh Division of Gastroenterology and Hepatology, Cedars-Sinai Medical Center, 8700 Beverly Blvd, Los Angeles, CA, USA |
| Reem Al Shabeeb | reem_shabeeb@gwmail.gwu.edu | MD | Department of Medicine, Inova Fairfax Medical Campus, Falls Church, Virginia, USA |
| Farah Ladak | Ladak@uthscsa.edu | MD | University of Texas Health, Department of Medicine, Division of Gastroenterology and Hepatology, San Antonio, Texas, USA |
| Jamie O. Yang | Joyang@mednet.ucla.edu | MD | University of California Los Angeles, Department of Internal Medicine, Los Angeles, CA, USA |
| Juan Echavarria | echavarriaj@uthscsa.edu | MD | University of Texas Health, Department of Medicine, Division of Gastroenterology and Hepatology, San Antonio, Texas, USA |
| Sumbal Babar | Sumbalbabar481@gmail.com | MBBS | University of Texas Health, Department of Medicine, Division of Infectious Diseases, San Antonio, Texas, USA |
| Aasma Shaukat | ashauka@gmail.com | MD | Division of Gastroenterology, Department of Medicine, NYU Grossman School of Medicine, New York, NY, USA |
| Samuel Margolis | 0009-0003-0374-2739; smargolis@mednet.ucla.edu | BS; Medical Student | David Geffen School Medicine University of California, Los Angeles, CA, USA |
| Nicholas P. Tatonetti | nicholas.tatonetti@cshs.org; 0000-0002-2700-2597 | PhD; Professor of Computational Biomedicine | 1- Department of Computational Biomedicine, Cedars-Sinai Medical Center, West Hollywood, California, USA; 2- Cedars-Sinai Cancer, Cedars-Sinai Medical Center, 8700 Beverly Blvd. Los Angeles, CA, USA; 3- Department of Biomedical Informatics, Columbia University, New York, New York, USA |
| Girish Nadkarni | girish.nadkarni@mountsinai.org | MD, MPH | 1- Division of Data-Driven and Digital Health (D3M), The Charles Bronfman Institute for Personalized Medicine, Icahn School of Medicine at Mount Sinai, New York, NY, USA; 2- The Charles Bronfman Institute of Personalized Medicine, Icahn School of Medicine at Mount Sinai, New York, NY, USA |
| Bara El Kurdi | baraabayern@gmail.com; 0000-0002-8048-7001 | MD | University of Texas Health, Department of Medicine, Division of Gastroenterology and Hepatology, San Antonio, Texas, USA |
| Ali Soroush | Ali.Soroush@mountsinai.org; 0000-0001-6900-5596 | MD, MS | 1- Division of Data-Driven and Digital Health (D3M), The Charles Bronfman Institute for Personalized Medicine, Icahn School of Medicine at Mount Sinai, New York, NY, USA; 2- The Charles Bronfman Institute of Personalized Medicine, Icahn School of Medicine at Mount Sinai, New York, NY, USA; 3- Henry D. Janowitz Division of Gastroenterology, Icahn School of Medicine at Mount Sinai, New York, New York, USA |

# Abstract

**Background and Aims:** This study evaluates the medical reasoning performance of large language models (LLMs) and vision language models (VLMs) in gastroenterology.

**Methods:** We used 300 gastroenterology board exam-style multiple-choice questions, 138 of which contain images to systematically assess the impact of model configurations and parameters and prompt engineering strategies utilizing GPT-3.5. Next, we assessed the performance of proprietary and open-source LLMs (versions), including GPT (3.5, 4, 4°, 4omini), Claude (3, 3.5), Gemini (1.0), Mistral, Llama (2, 3, 3.1), Mixtral, and Phi (3), across different interfaces (web and API), computing environments (cloud and local), and model precisions (with and without quantization). Finally, we assessed accuracy using a semiautomated pipeline.

**Results:** Among the proprietary models, GPT-4o (73.7%) and Claude3.5-Sonnet (74.0%) achieved the highest accuracy, outperforming the top open-source models: Llama3.1-405b (64%), Llama3.1-70b (58.3%), and Mixtral-8x7b (54.3%). Among the quantized open-source models, the 6-bit quantized Phi3-14b (48.7%) performed best. The scores of the quantized models were comparable to those of the full-precision models Llama2-7b, Llama2-13b, and Gemma2-9b. Notably, VLM performance on image-containing questions did not improve when the images were provided and worsened when LLM-generated captions were provided. In contrast, a 10% increase in accuracy was observed when images were accompanied by one-sentence human-crafted image descriptions.

**Conclusion:** In conclusion, while LLMs exhibit robust zero-shot performance in medical reasoning, the integration of visual data remains a challenge for VLMs. Effective deployment involves carefully determining optimal model configurations, encouraging users to consider either the high performance of proprietary models or the flexible adaptability of open-source models.

**Keywords:** Large Language Models, Vision Language Models, Gastroenterology, Medical Reasoning, Quantitated Models

# Introduction

Large language models (LLMs) are capable of processing and generating large amounts of text with a high degree of sophistication. An intriguing application of LLMs is assisting physicians by providing decision support, either through their own processing capabilities or by leveraging evidence-based knowledge [1], [2]. However, the performance of LLMs varies significantly across different tasks and domains [3]. For example, the models can perform well at summarizing and extracting data from clinical documents but struggle with assigning billing codes to those documents [4], [5]. Before LLMs can be safely applied to critical clinical tasks, their capacity for accurate, safe, and reliable decision-making must be comprehensively assessed [6].

The evaluation of specialty-specific performance is particularly important, given the limited availability of training data and the increased complexity of reasoning required in specialized medical fields. Moreover, each medical specialty presents unique challenges and considerations. In the field of gastroenterology, for example, diagnostics require the integrated analysis of medical imaging, clinical narratives, and tabular data. These diverse data types can potentially be processed via vision-language models (VLMs), which are designed to handle both textual and visual information [7]. However, the efficacy of VLMs in this specific context warrants further investigation.

Multiple-choice questions (MCQs) serve as common benchmarks of LLM medical reasoning capabilities [6]. While MCQs can provide an objective and concrete gold standard to benchmark against, they do not capture the full complexity of real-world clinical reasoning [8], [9]. Nevertheless, LLMs such as GPT-3.5, GPT-4, and Bard have achieved varying levels of accuracy on MCQs for many different medical specialties. In some cases, they match the performance of medical residents (**Table 1**) [10], [11], [12], [13], [14], [15], [16], [17], [18], [19], [20], [21], [22], [23], [24]. These findings must be viewed with caution, however, as there is significant methodological heterogeneity across the studies. In addition, there is a lack of specialty-level analysis of medical reasoning for VLMs.

**Table 1: LLMs and human performance on specialty and subspecialty board exams.**

| Field; Author | Dataset | Handling Image-Inclusive Qs | Evaluation Metrics | Stratified Performance | Reporting Prompt (Prompt) | Reporting Usage Method | Reporting LLM Version (Version) | Reporting Evaluation | LLM-Performance | Human Performance |
|---|---|---|---|---|---|---|---|---|---|---|
| Soft Skil [24] | 80 soft skill Qs USMLE SA (AMBOSS & NBME) | Excluded | • Accuracy<br>• Consistency | None | Sufficient (QA followed by "are you sure?") | Sufficient | Insufficient | Insufficient | • GPT4: 90%<br>• GPT3.5: 62.5% | • Average: 78% |
| USMLE [10] | 100 step 1 USMLE SA (AMBOSS & NBME) | Excluded | • Accuracy<br>• Error source (logic, internal, or external information) | None | Sufficient (Just QA) | Sufficient | Sufficient (Dec 15 version) | Sufficient | • GPT3.5: 56%, 59%, 44%, 42%<br>• InstructGPT: 45%, 54%, 36%, 35%<br>• GPT3: 22%, 19%, 29%, 17% | • Pass: 60% |
| USMLE [11] | 367 step 1 USMLE SA (USMLE 2022 SA) | Excluded | • Accuracy<br>• Concordance<br>• Insight | None | Sufficient (three variants: open-ended, MCQ without justification, MCQ with justification) | Insufficient | Insufficient | Sufficient | • GPT3.5: Step 1, 41.2%; Step 2, 49.5%; Step 3, 59.8% | • Pass: 60% |
| Medicine | 198 residency entrance exams | Excluded | • Accuracy | • Subcategory<br>• Language | Sufficient (Just QA) | Partially Sufficient | Sufficient (May 2023) | Insufficient | GPT4:<br>  Persian: 81.3%<br>  English: 84.3%<br>  French: 81.8%<br>  Spanish: 83.8% | • Average: 58.3%<br>• Top 2%: 83.3% |
| Ophthalmology [12] | 467 medicine SA (StatPearls) | Images replaced by image description | • Accuracy | • Difficulty<br>• Subcategory | Sufficient (Just QA) | Sufficient | Sufficient (May 2023) | Insufficient | • GPT4: 73.2%<br>• GPT3.5: 55.6% | • Average: 58.1% (****) |
| Ophthalmology [13] | 125 board SA (OphtoQuestion) | Excluded | • Accuracy | • Subcategory<br>• Word length | Sufficient (Just QA) | Sufficient | Insufficient (no model name) | Partially Sufficient | • GPT3.5: 46.4% | NA |
| Ophthalmology[14] | 250 board SA | Answered without image information | • Accuracy<br>• Hallucination | • Patient Care Phase<br>• Difficulty | Sufficient (Just QA) | Insufficient | Sufficient (March 2023) | Insufficient | • GPT3.5: 58.8%<br>• GPT4: 71.6% | • Average: 72.2% |
| Gastroenterology [15] | 455 board SA (ACG) | Excluded | • Accuracy | • Subcategory<br>• Difficulty | Sufficient (Just QA) | Insufficient | Sufficient (March 2023) | Insufficient | • GPT3.5: 65.1%<br>• GPT4: 62.4% | • Pass: 70% |

| Field; Author | Dataset | Handling Image-Inclusive Qs | Evaluation Metrics | Stratified Performance | Reporting Prompt (Prompt) | Reporting Usage Method | Reporting LLM Version (Version) | Reporting Evaluation | LLM-Performance | Human Performance |
|---|---|---|---|---|---|---|---|---|---|---|
| Nephrology [16] | 99 board SA in Japanese language | Answered without image information | • Accuracy | • Taxonomy<br>• Subcategory<br>• image-inclusive<br>• Year | Sufficient (QA + "answer following question") | Sufficient | Insufficient | Insufficient | • GPT3.5: 31.3%<br>• GPT4: 54.5%<br>• Bard: 32.3% | • 1st year avg: 36.4%<br>• 3nd year avg: 49.5%<br>• 4th year avg: 67.7% |
| Cardiology [17] | 362 board exam (EECC) | Excluded | • Accuracy | - | Sufficient (Just QA) | Partially Sufficient | Insufficient | Partially Sufficient | • GPT3.5: 58.8% | • Pass: ~60% |
| Dermatology [18] | Specialty certificate exam | Excluded | • Accuracy | • Subcategory | Sufficient (Just QA) | Insufficient | Insufficient | Insufficient | • GPT3.5: 63%<br>• GPT4: 90% | • Pass: ~70% |
| Plastic Surgery [19] | 1129 in-service exam | Excluded | • Accuracy | • Year<br>• Subcategory | Sufficient (Just QA) | Sufficient | Insufficient | Partially Sufficient | • GPT3.5: 55.8% | • 1st year: 49%tile<br>• 2nd year: 13%tile<br>• 3rd year: 5%tile |
| Otolaryngology [20] | 479 MCQ and 2097 single-choice Board SA | Excluded | • Accuracy | • Subcategory | Sufficient (Please answer the following question. Note that only one option is correct: + QA) | Partially Sufficient | Sufficient (3 May version) | Insufficient | •GPT3.5: 34% on multiple-choice and 57% on single-choice | NA |
| Neurosurgery [22] | 149 board SA (SANE) | Answered without image information | • Accuracy<br>• Hallucination | • Subcategory<br>• Complexity<br>• Word length | Sufficient (Just QA) | Insufficient | Sufficient (March and April 2023) | Insufficient | • GPT3.5: 62.4%<br>• GPT4: 82.6%<br>• Bard: 44.2% | NA |
| Orthopedic Surgery [21] | 212 in-training exam(OITE) | Excluded | • Accuracy | • Taxonomy | Sufficient (Q + "Select the single best answer" + Options) | Partially Sufficient | Insufficient | Sufficient | • GPT3.5: 47% | • Pass: 10th percentile (GPT failed)<br>• PGY1: 40th %tile |

Footnote: Abbreviations: SA, self-assessment; Q, question; OITE, Orthopedic In-Training Examination; EECC, European Exam in Core Cardiology; SANS, Self-Assessment Neurosurgery Examination; MCQ: multiple choice question, ****: statistically significant compared with the LLM model

This study aims to address these gaps by providing a detailed and systematic evaluation of the performance of both LLMs and VLMs to show the extent of their medical reasoning. We explore their ability to interpret complex gastroenterology-specific medical images and text, aiming to establish reproducible methods and benchmarking strategies that can be extended to other medical MCQ benchmarks. We assess open- and closed-source models, cloud and local computing environments, and quantized open-source models. A list of glossaries and terms related to LLMs is provided in **Supplementary Table S1**.

## Methods

### Dataset characteristics

To assess gastroenterology reasoning, we utilized gastroenterology board exam style questions from the American College of Gastroenterology (ACG) self-assessment examinations. Self-assessments are meant to represent the knowledge, skills and attitudes required for excellent patient care in gastroenterology. Gastroenterologists The educational pathway for participants, including established gastroenterologists and fellows, comprises eight years of medical education to earn a Doctor of Medicine (MD) degree, followed by six years of specialty and subspecialty training. The full details of the exam context are available in **Supplementary Section 1**.

We used the 2021, 2022, and 2023 editions of the ACG self-assessments. Each edition consisted of 300 questions with three to five answer choices and a single best answer. There were 124, 138, and 128 questions with images from the 2021, 2022, and 2023 exams, respectively. Each image question contained between one and four endoscopy, radiology, histology, or physical examination images. Access to these exams and answers is restricted by a paywall, which presumably prevents their inclusion in LLM training datasets.

### Experimental Design: Assessing LLM Reasoning for Gastroenterology MCQs

We structured our analysis of LLM performance in terms of gastroenterology reasoning into several experiments, which are briefly summarized in **Figure 1**. In Experiment 0, we used a subset of our dataset and GPT-3.5 to determine the best performing model settings (function, prompt, temperature, and max-token) for our subsequent experiments (**Supplementary Section 2**). In Experiment 1, we evaluated the performance of all LLMs on both 162 text-only and 138 image-inclusive questions in the 2022 ACG self-assessment. (**Supplementary Section 3**). In Experiment 2, we assessed the performance of the VLMs on the 138 image-containing questions from the 2022 ACG self-assessment, comparing performance at baseline with performance with LLM-generated captions or human-generated captions. (**Supplementary Section 4**). We evaluated the historic trends of GPTs to analyze the impact of model version on model performance and assess indirect evidence of data leakage in our dataset.

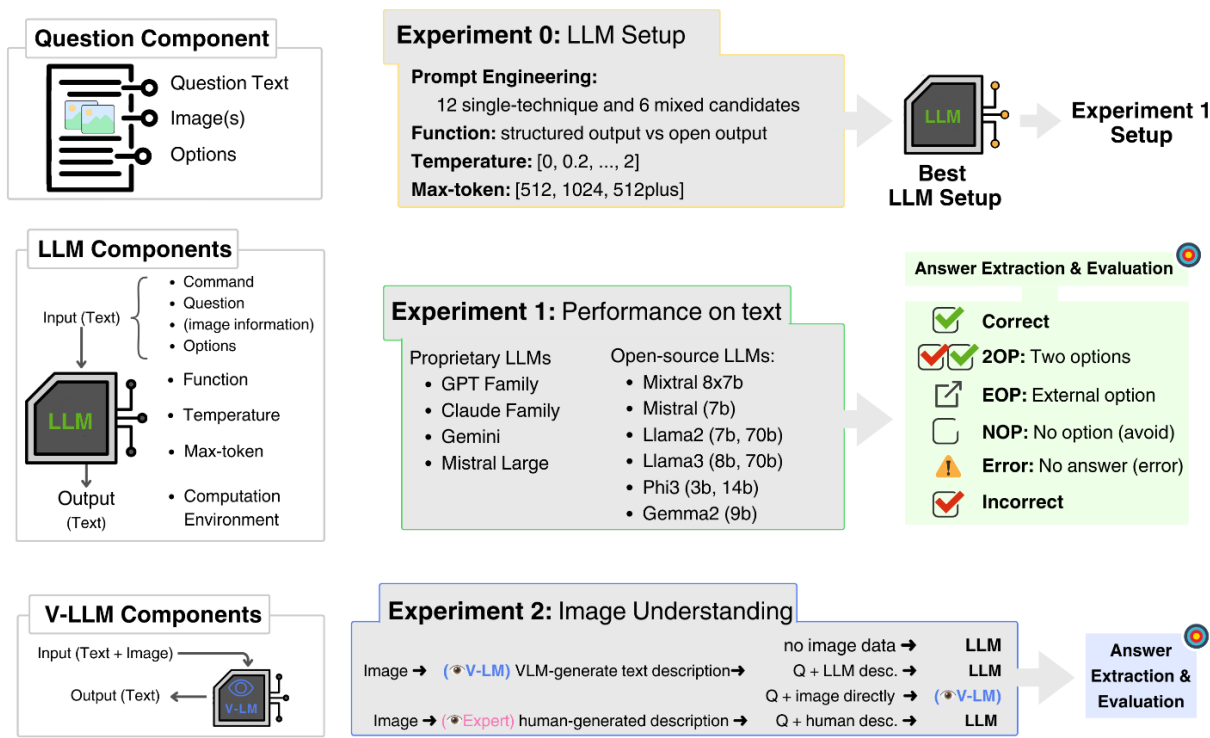

**Figure 1. Summary of the data and LLM components, along with the three experimental designs.**
*Image Description: The right column shows the components of multiple-choice questions, large language models (LLMs), and vision language models (VLMs). Experiment 0 aims to find optimized model parameters, including the input text itself, function (*i.e.*, the method used to provide input and obtain output), max-token (the total number of output tokens, including the input length), and temperature (which determines the randomness in the output text). The models in Experiment 1 and the labels used during evaluation are highlighted in green. The blue box illustrates the four scenarios of providing questions along with (1) no image data, (2) LLM-generated descriptions, (3) images directly, and (4) human-generated descriptions.*

## Experiment 0: LLM Setting Optimization

To identify the optimal settings for our analysis of LLM reasoning performance, we systematically evaluated the impacts of temperature, max_token (maximum input and output token count), prompt design, and output structure. We used a sample of 60 questions from ACG 2021 and 2022 as the optimization dataset and selected the GPT-3.5 API for its efficiency and usability. We assessed six temperature settings, three max_token settings, eleven prompt engineering strategies, and two data output structures via the GPT-3.5 API. Each scenario was replicated three times, and comparisons were made to find the best value of each parameter, step-by-step. This approach allowed us to first identify the best function, followed by prompt, max_token, and temperature. Detailed information is provided in **Supplementary Sections 2.1 and 2.2**.

Our analysis identified the following optimal model settings and inputs: (1) temperature of 1, (2) max_token equal to the number of input tokens plus 512 output tokens, (3) use of a function call to generate a structured output, and (4) the following prompt: "*Imagine you are a seasoned gastroenterologist approaching this complex question from the gastroenterology board exam. Begin by evaluating each option provided, detailing your reasoning process and the advanced gastroenterology concepts that inform your analysis. Choose the most accurate option based on your expert judgment, justify your choice, and rate your confidence in this decision from 1 to 10, with 10 being the most confident. Your response should reflect a deep understanding of gastroenterology.*" We applied these settings when they were available. Otherwise, we used the default settings (see **Supplementary Section 2.3**).

## Experiment 1: Text-Based Gastroenterology Questions

We assessed the performance of LLMs on text-based MCQs in different environments. In this study, "environments" refer to the methods used to input data into and receive outputs from LLMs, which are categorized into web interfaces, Poe interfaces, API (cloud computing), and local computer setups. We utilized the first-party web interfaces OpenAI and Claude; the third-party web interfaces Poe, OpenAI and Claude API; and a local computer setup. API-based experiments were conducted in Python, providing more control over model settings such as prompt, temperature, max_token, seed, and function tools (details in **Supplementary Table 1**). A broad range of LLMs were assessed, as listed in **Table 2**. The exact names of the models used, their usage dates, and the sources of the local models are detailed in **Supplementary Section 3.1**.

**Table 2. List of large language models used in experiment 1** (exact model names and times of use are mentioned in Supplementary Section 3)

| Proprietary Web Chat Interface | Model Setup (Parameters) | Proprietary Web-based API | Model Setup (Parameters) |
|---|---|---|---|
| GPT-3.5 Web | P | GPT-3.5 API | P, T, MT, S |
| GPT-4 Web | P | GPT-4 API | P, T, MT, S |
| -- | | GPT-4o API | P, T, MT, S |
| -- | | GPT-4omini API | P, T, MT, S |
| Claude3-Haiku Web | P | Claude3-Haiku API | P, T, MT |
| Claude3-Sonnet Web | P | Claude3-Sonnet API | P, T, MT |
| Claude3-Opus Web | P | Claude3-Opus API | P, T, MT |
| -- | | Claude3.5-Sonnet API | P, T, MT |
| Gemini Web | P | -- | |
| Gemini Advanced Web | P | -- | |
| **Open-Source Web-based Poe Interface** | | **Open-Source Local API** | |
| Llama2-7b Poe | P | Llama2-7b-Q8 | P, T, MT, GPU layers |
| Llama2-13b | P | Llama2-13b-Q5KM | P, T, MT, GPU layers |
| -- | | MedicineLLM-Q8 | P, T, MT, GPU layers |
| Llama2-70b | | -- | |
| Llama3-8b Poe | P | Llama3-8b-Q8 | P, T, MT, GPU layers |
| Llama3-70b Poe | P | OpenBioLLM-Q8 | P, T, MT, GPU layers |
| Llama3.1-8b Poe | P | -- | |
| Llama3.1-70b Poe | P | -- | |
| Llama3.1-405b Poe | P | -- | |
| Llama3-70b Poe | P | -- | |
| Mistral-7b | P | Mistral-7b-Q8 | P, T, MT, GPU layers |
| Mistral-Large Poe | P | -- | |
| Mixtral-8x7b Poe | P | -- | |
| Gemma-2-9b Poe | P | Gemma2-9b-Q8 | P, T, MT, GPU layers |
| Gemma-2-27b Poe | P | -- | |
| -- | | Phi3mini-3b-fp16 | P, T, MT, GPU layers |
| -- | | Phi3medium-14b-Q6 | P, T, MT, GPU layers |

*Footnote: Parameter abbreviations: P, prompt, T, temperature, MT, max-token; GPU layers, the number of layers offloaded to the GPU*

## Experiment 2: Image-Based Gastroenterology Questions

For experiment 2, we assessed the visual reasoning capabilities of proprietary VLMs via image-containing questions from the 2022 ACG self-assessment. We interact with the web interfaces of GPT-4, Claude3-Opus, and Gemini Advanced and the API interfaces of GPT-4 and Claude3-Opus. We assessed performance across four distinct scenarios to capture different aspects of visual understanding.

- (1) No-image: Excluding the image and providing only the text to the VLM, which establishes the baseline performance without image information. This scenario reflects the base performance without visual context.

- (2) Human-generated description: This uses human-crafted hints that point to the information in lieu of the actual images and guide toward the correct answer. This scenario represents the maximum performance when the correct information embedded within the image is available.
- (3) LLM-generated description: First, captions for images are generated via VLMs; then, this description is placed after the question stem and before the options are provided, and it is provided to the same VLM.
- (4) Direct image: Directly providing the image along with the question text.

The full prompts for the scenarios are detailed in **Supplementary Section 4.1**.

**Experiment 3: Trend of Performance**

We performed additional sensitivity analyses to assess the impacts of model updates and training date cutoffs on performance. To evaluate the impacts of model updates on model performance, we examined the OpenAI model family, as this provided the largest number of model versions to compare. compared performance on text-based questions from the 2022 ACG-SA for all available OpenAI LLMs. This included GPT-3 (babbage-002, davinci-002), GPT-3.5 Turbo (gpt-3.5-turbo-1106, gpt-3.5-turbo-0125), GPT-4 (gpt-4-0613), GPT-4 Turbo (gpt-4-1106-preview, gpt-4-0125-preview), and GPT-4o (gpt-4o-2024-05-13). Historical models were only available via the web-based API.

**Auxiliary Results: Consistency and Cost**

We conducted an auxiliary investigation of the consistency and expectation cost and time of the model. To investigate the consistency of the model outputs, we conducted setup experiments with multiple runs via GPT-3.5. We measured the variations in model accuracy across three separate runs. Specifically, we examined the effects of using a fixed random seed, different temperatures, and prompt complexity on the consistency of the model outputs.

To estimate the cost and runtime of the models, we analyzed a median question length of 216 tokens. For a detailed case study, a medium-difficulty question with 217 tokens was selected, with a focus on liver disease (using the same settings as in Experiment 1). After a prompt-engineered command was applied, the total input increased to 311 tokens. The execution time was measured from the input to the final token via the time library when possible or a timer for web interfaces.

Token counts were managed via tiktoken, with pricing data from OpenAI and Anthropic as of June 20, 2024. Paid accounts for OpenAI, Claude3, and Gemini services and costs $20/month. The cost per token was retrieved from the official websites of OpenAI and Claude3 for models run in the API environment.

**Evaluation of LLM Responses**

We used a semiautomated approach to evaluate LLM answers, briefly summarized in **Figure 2**. LLM responses can be structured (i.e., the output is a dictionary-like object) or unstructured (i.e., raw textual response). For structured responses, we compared LLM responses against the ACG self-assessment answer key. A correct response was defined as the selection of a single, correct answer, whereas an incorrect response was the selection of a single, incorrect answer. For the remaining models and environments with unstructured responses, we used the GPT-3.5 API to extract the selected option from the LLM textual response. The GPT-3.5 option was either the extraction of an option or the labeling of the question to be manually evaluated by a human. The human annotator (SAASN) then further classified the response into the final categories:

- Correct: When the LLM chooses one option that is correct.
- Incorrect: When the LLM chooses one or two options that are incorrect.
- 2OP: When the LLM chooses two options, one of which is correct.
- EOP: When the LLM selects an external option or answer.
- NOP: When the LLM avoids providing an option due to a lack of sufficient information.
- Error: When the LLM's answer was incomplete or nonsensical due to a technical error after two tries.

Further details of the evaluation process are available in **Supplementary Section 5** and **Supplementary Figure S1**. We randomly selected a dataset of 100 responses generated through various environments and LLMs for validation of our evaluation approach. A human reviewer, blinded to the extraction label, reviewed the questions and provided labels. In 13% of the cases, human validation was necessary. Among the 87 cases where GPT-3.5 identified an option in the response, one incorrect option was evident, resulting in an overall accuracy of 99% for our approach (**Supplementary Figure S2).** This approach reduced our evaluation tasks from 11,100 question-answers to 451 pairs.

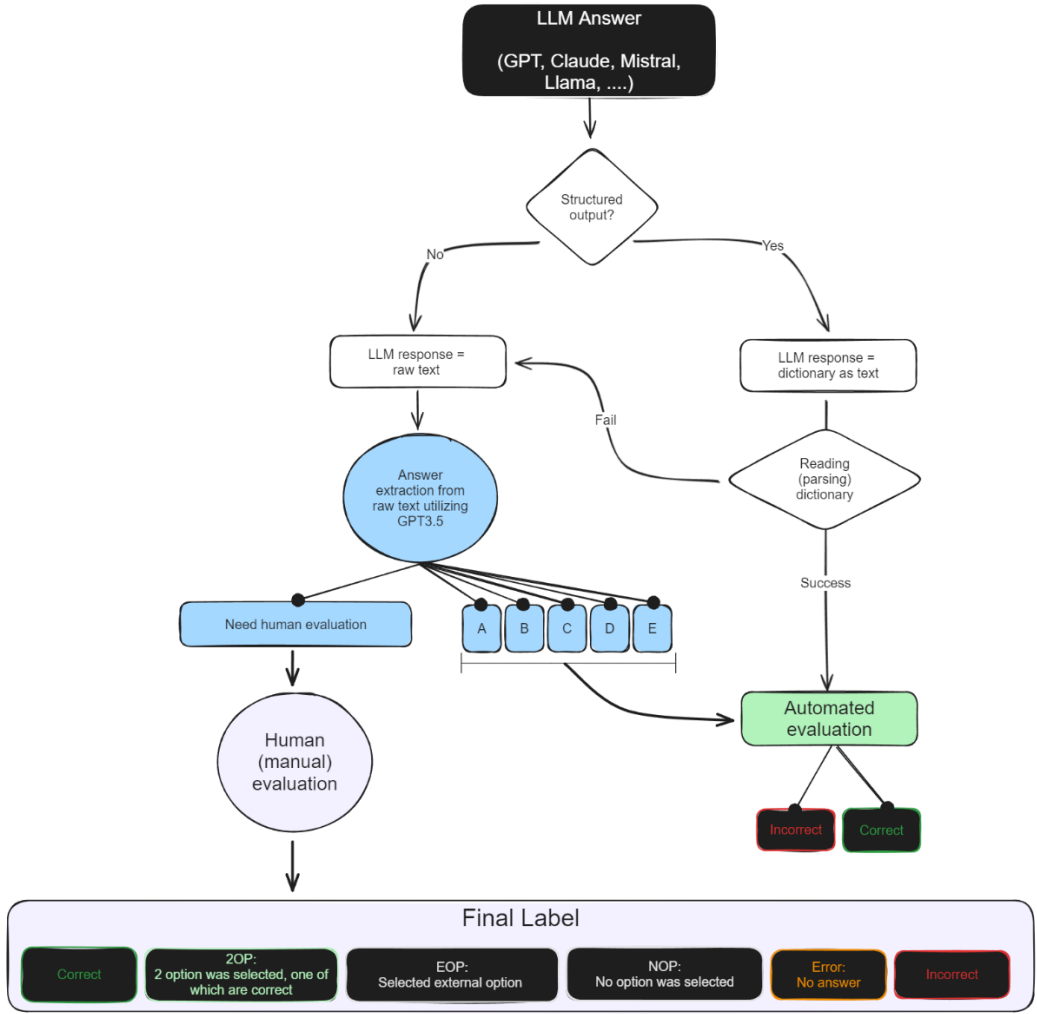

**Figure 2. Semiautomated evaluation pipeline for LLM responses**

*Image Description: We used a semiautomated approach to evaluate LLM answers. LLM responses can be structured (i.e., the output is a dictionary-like object) or unstructured (i.e., raw textual response). To evaluate the structured outputs, we read (parsed) the dictionary, retrieved the selected option, and compared it to the correct option for evaluation. For all other models in web interfaces, Poe, or local computer environments with unstructured responses, we used GPT-3.5 to extract the selected option from the raw textual response. For uncertain labels marked by GPT-3.5, a human evaluator assessed the response. The final labels were as follows: correct, when the LLM chose one option that was correct; incorrect, when the LLM chose one or two options that were incorrect; 2OP, when the LLM chose two options that were correct; EOP, when the LLM selected an external option or answer; NOP, when the LLM avoided providing options due to a lack of sufficient information; and Error, when the LLM answer was incomplete or nonsense due to a technical error after two tries.*

**Performance stratified by question type**

We reported the performance stratified by question type to identify areas where the LLMs' answers were less accurate. The full details are available in **Supplementary Section 5.3 Questions.** On the basis of the percentage of test-takers who answered a question correctly, we classified the questions into four difficulty quartiles. We used simplified Bloom's Taxonomy to classify questions on the basis of the cognitive level tested (lower-order, higher-order, case-based, and integrated). The question phase of care was defined as either diagnosis**,** treatment, management of complications, or pathophysiology. The question topic categories were provided by the ACG. The question token length was determined via the tiktoken Python library and categorized into tertiles of short (49-179 tokens), medium (180-262 tokens), and long (263-588 tokens), which resulted in 99, 99, and 102 labels, respectively. The full details are available in **Supplementary Section 5.3 Questions Classification: Difficulty, Taxonomy, Care Phase, Subject, and Length**.

## Libraries and Local Computing

We used Python version 3.11 for API interactions, and we used Langchain (versions 0.1.17 and 0.2.4) to standardize these interactions. To classify and troubleshoot LLM responses, we created an app for human evaluation via the Streamlit library. Communication with the OpenAI models was performed via the OpenAI Python package. To interact with the Claude3 models, we provided XML-formatted requests to the API. Local models were run on a laptop equipped with an RTX 3080 Ti GPU (16 GB VRAM) and a Core i9-12900 CPU (32 GB DDR5 RAM), creating a local server utilizing the llama.cpp library and LM Studio version 0.2. These runs utilized half of the model layers on the GPU (typically 16) and 10 CPU threads. The tiktoken library was employed to count the number of tokens in the question stem and options, particularly in scenarios using a 512-plus max-token limit. We used asyncio for sending parallel requests whenever possible.

## Statistical analysis

Categorical variables are presented as counts (N) and percentages (%). Continuous variables are presented as the means ± standard deviations (SDs). Additionally, we reported the median and min–max range formatted as the median [range: lower range, higher range]. For comparison of prompt engineering techniques, we used the Wilcoxon rank test to calculate the P value using the "raw prompt" as the reference, and the 95% confidence interval (95% CI) was estimated via the bootstrapping method with 10000 samples. For the comparison of the four scenarios in Experiment 2, we used a paired t test with "no image" as the reference. Python was used for statistical analysis and visualization, employing the pandas, matplotlib, scipy, and statsmodels libraries. Where relevant, p values were two-tailed, and a p value less than 0.05 was considered to indicate statistical significance.

**Ethical considerations**

This study employed multiple-choice questions from GI board examinations, emphasizing the ethical use of data. As the study did not involve human subjects, formal approval was not needed. We placed a strong emphasis on accurately citing and attributing the questions to their original sources to maintain academic integrity and respect intellectual property rights. Additionally, we ensured that data usage was terminated in services such as OpenAI, Claude3, and Poe to prevent unintended use of the data by these platforms [25], [26], [27].

# Results

## Dataset characterization

The Experiment 1 dataset, ACG-SA 2022, comprises 300 questions, 138 of which are image-based. The dataset spans 10 gastroenterology topics, with the five most prevalent being liver (N=52), colon (N=49), esophagus (N=36), pancreaticobiliary (N=32), and endoscopy (N=26). Nearly all the questions were case-based (N=297) and targeted higher-order thinking skills (N=298). We identified 123 questions focused on diagnosis, 217 on treatment, 211 on investigation, 55 on complications, and 3 on pathophysiology. The average question-answer length was 232.92 ± 94.10 tokens, roughly equivalent to 174 words. The average performance of test-takers on 2022 questions was 74.52% ± 19.49%.

The optimization dataset consisted of 60 randomly sampled text-based questions from ACG-SA 2021 and 2023. The Experiment 2 dataset comprises 138 image-based questions from ACG-SA 2022, totaling 195 images used to answer these questions. The average dimensions of the images were 257 ± 79 pixels by 206 ± 83 pixels, with a resolution of 110 ± 71 DPI. The Experiment 3 dataset included ACG-SA 2021, 2022, and 2023, each containing 300 questions.

## LLM Setting Optimization

We found that the choice of model settings, prompt strategy, and use of structured outputs all impacted LLM performance in gastroenterology MCQ reasoning. We found that structured output generation, when available, enhanced performance by 5–10% (**Figure 3.a and**

**Supplementary Table S2**). When evaluating the 11 prompt engineering techniques individually, we found that 5 techniques resulted in a score > 2 standard deviations above the raw prompt. The *direct question and answering and justification* prompts had the greatest impact on performance (+4.44%), followed by *contextual embedding* (+2.77%), *confidence scoring* (+1.67%), and *expert mimicry* (+1.67%). The best performing prompt containing all 5 techniques increased performance by 5.00% [95% CI: 1.11%, 7.78%] (**Figure 3.d and Supplementary Table S3**). Setting the model temperature to 1 (default) resulted in optimal performance (**Figure 3.b and Supplementary Table S4**). Additionally, setting max_token to 512 plus the number of tokens in the input text resulted in the best performance (**Figure 3.c** and **Supplementary Table S4**). The best selected settings were used for all subsequent analyses.

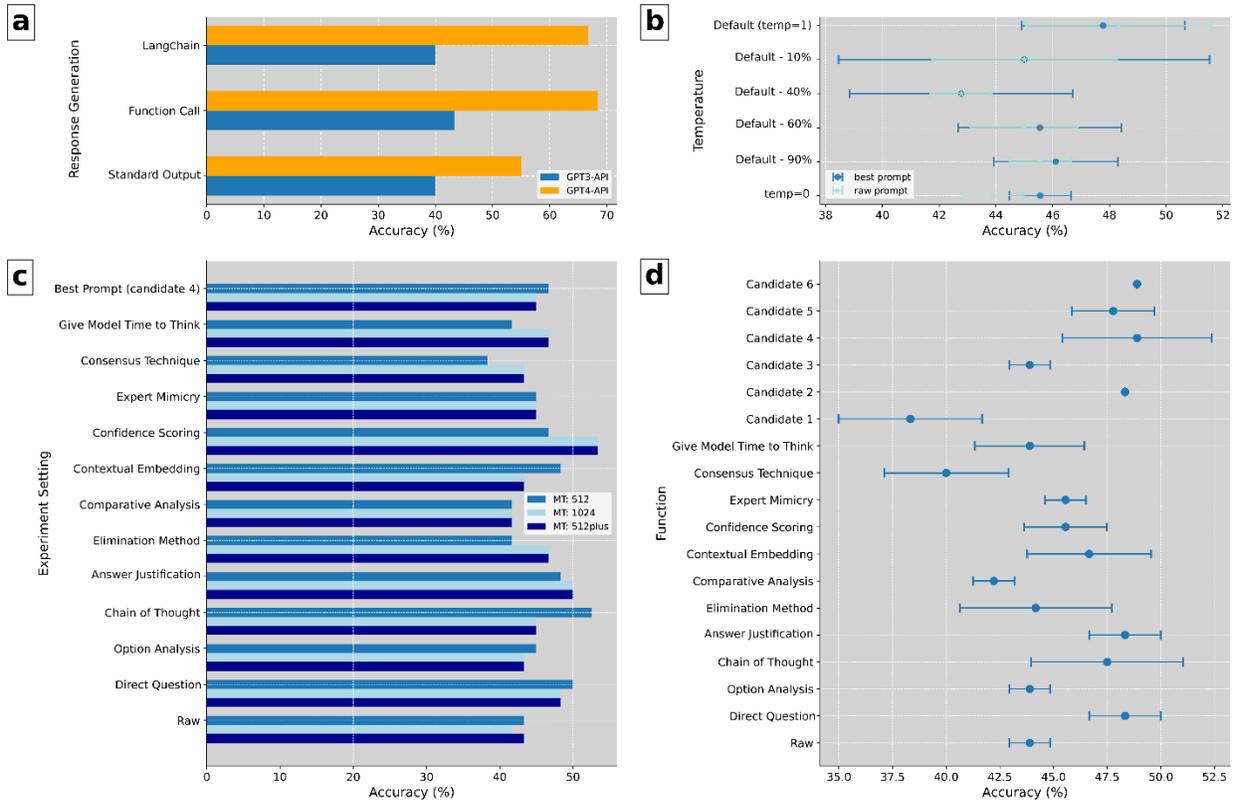

**Figure 3. Impact of** output structure (a), temperature **(b), Max_token (c), and** prompt engineering **(d) on LLM MCQ** performance.

*Figure Legends: (a) Accuracy comparison of different structured output strategies (LangChain, Function Call, Standard Output). (b) Accuracy results at various temperature settings, highlighting the optimal default setting and the consistency observed at lower temperatures. (c) Accuracy of various max-token lengths (512 tokens, 1024 tokens, and input token length plus 512 tokens), illustrating the benefits of fair token allocation across questions. (d) Accuracy (average and 95% CI) of different individual and combined (candidate) prompt engineering techniques.*

## LLM Performance on Text-based MCQ Reasoning

**Figure 4** illustrates the performance for all 2022 ACG self-assessment questions when only the question textual information was provided. As depicted in **Supplementary Figure S3**, the granular analysis at the question level reveals a significant overlap in correctly answered questions across various LLMs. GPT-4o achieved the highest score of 73.7%, closely matching the average score of test takers (74.52%) and surpassing the 70% cutoff required to pass the exam. When restricted to only the questions without associated image data, the GPT-°performance increased to 75.9%.

Among the remaining commercial models, GPT-4 and Claude 3-Opus scored 66.0%, followed by Mistral-Large with 60.3% on the entire 2022 ACG self-assessment. Notably, the performance of the API interface with optimized model settings was uniformly equivalent to the performance of the web chat interface for all proprietary models, suggesting that unoptimized API use may be inferior to the web interface for these models.

Among the open-source models, Llama3.1-405b scored best (64%), followed by Llama3.1-70b and Mixtral-8x7b (58.3% and 54.3%, respectively). The best locally hosted quantized model was Phi3-14b-Q6, which correctly answered 48.7% of the questions. A comparison of the full-precision open-source models with their quantized counterparts revealed comparable results for 3 out of the 4 models. The performance change was greatest for Llama3-8b, which increased from 43.3% to 31% accuracy after 8-bit quantization. In contrast, the full-precision and 8-bit quantized versions had similar but overall poor accuracies, considering Llama2-7b (30.7% vs. 30.3%), Mistralv2-7b (40.7% and 37.3%), and Gemma2-9b (44.7% vs. 45.3%).

We also tested two fine-tuned versions of the Llama model family that were 8-bit quantized to assess the impact of model fine-tuning: medicineLLM (Llama2-7b) and OpenBioLLM (Llama3-8b). Both models performed worse than their source models, with OpenBioLLM performing worse than Llama3-8b (29% vs. 31%) and medicineLLM performing worse than Llama2-7b (27% vs. 30.3%).

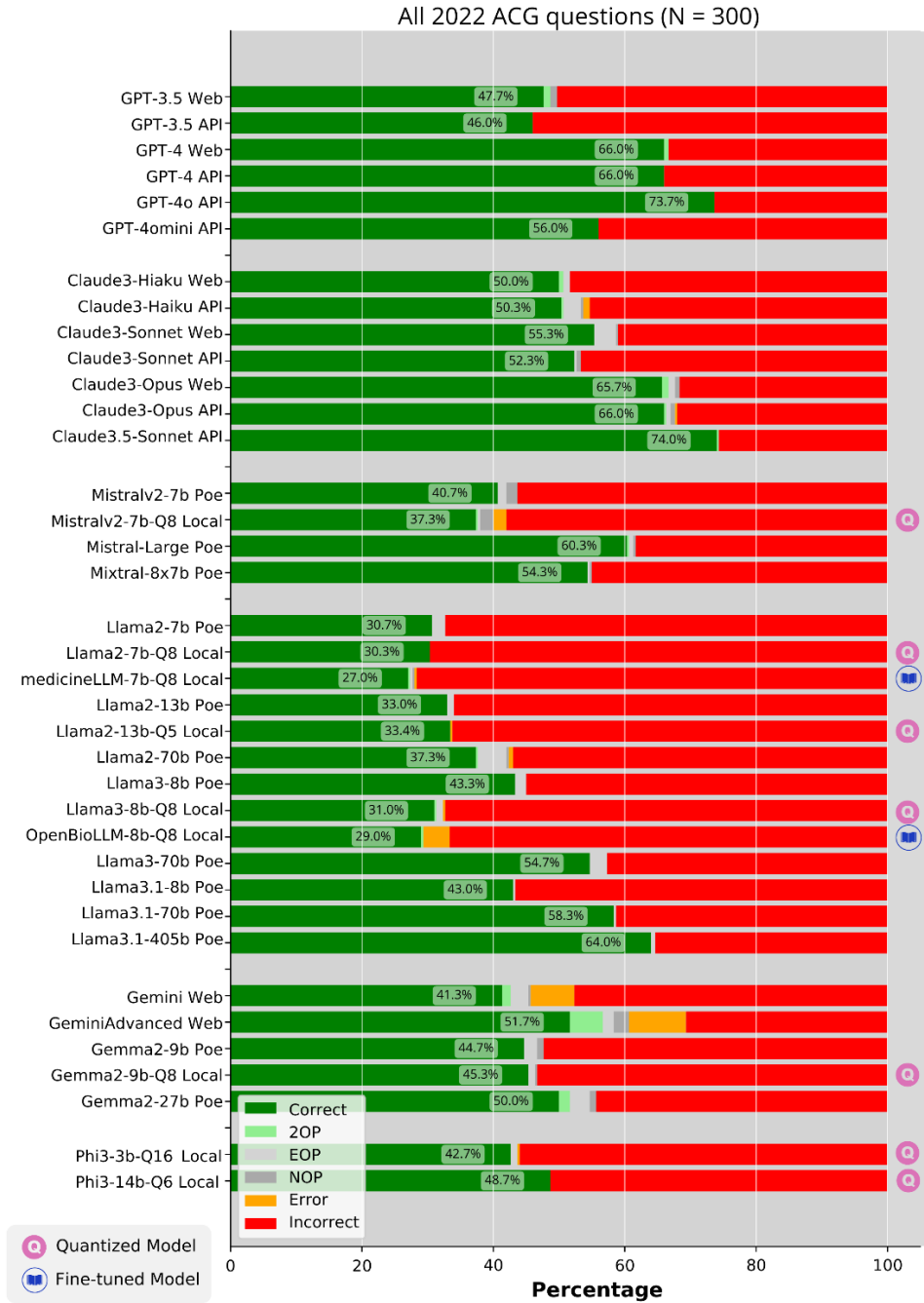

**Figure 4. LLM Performance on Text-Only Gastroenterology Multiple-choice Questions.**
*Image Description: The figure shows the performance of large language models (LLMs) on Gastroenterology Board Exam questions. The bars are clustered considering model families (architecture) and sorted from oldest to the newest in the clusters. The quantified models and models fine-tuned on medical data are highlighted with pink and blue icons, respectively.*
*2OP means that two options were chosen, one of which is the correct option. EOP means that an external option was chosen that was not included in the provided options. NOP means that no option was selected by LLM due to the need for caution. Error means that no rational answer was provided due to failure in generation or an unrelated response after two tries.*

**Performance Stratified by Question Type**

We examined question characteristics to assess their impact on LLM performance. There were no notable performance differences based on image-based vs. text-based questions (**Supplementary Figure S4**), content subjects (**Supplementary Figure S5)** or patient care phases (**Supplementary Figure S6).** In contrast, performance decreased with increasing question difficulty across all LLMs (**Supplementary Figure S7**). For example, GPT-4o's performance decreased from 88% on easy questions to 74% on medium difficulty questions and 58.7% on challenging questions. When the impact of question length was examined, performance interestingly improved with longer questions (**Supplementary Figure S8**). Since only two questions were aimed at lower-level thinking skills and three questions were not case-based, we did not investigate performance stratified by question taxonomy.

**VLM Performance on Image-Inclusive Questions**

Compared with scenarios without images, providing images directly to VLMs did not improve performance. This was observed for GPT4V Web (mean difference, 95% CI: -4.87%, 10.21%), GPT4V API (mean difference, 95% CI: -11.18%, 3.84%), Claude3-OpusV Web (mean difference, 95% CI: -7.7%, 7.7%), Claude3-OpusV API (mean difference, 95% CI: -6.43%, 8.43%), and GeminiAdvancedV Web (mean difference, 95% CI: -11.36%, 2.02%). Similarly, generating VLM descriptions and then supplying the descriptions along with the question text did not increase performance. This approach did not improve the performance of GPT4V or GeminiAdvancedV and decreased the performance of Claude3-OpusV by -15.2% (mean difference, 95% CI: -22.85%, -7.15%).

In contrast, providing a one-sentence human hint that captured the core information from the image resulted in performance improvements for all the models. These findings suggest that LLM gastroenterology image comprehension and reasoning are poor. Human image summaries resulted in improvements ranging from 8.0% (95% CI: 0.65%, 15.35%) for GPT4V Web to 29.3% (95% CI: 21.86%, 36.8%) for GeminiAdvancedV Web. The only exception was the GPT4V API, which showed no statistically significant improvement. **Figure 5** illustrates the performance of VLMs in answering multiple-choice questions across four different scenarios: no image data, LLM-generated text descriptions, images, and human-generated text descriptions.

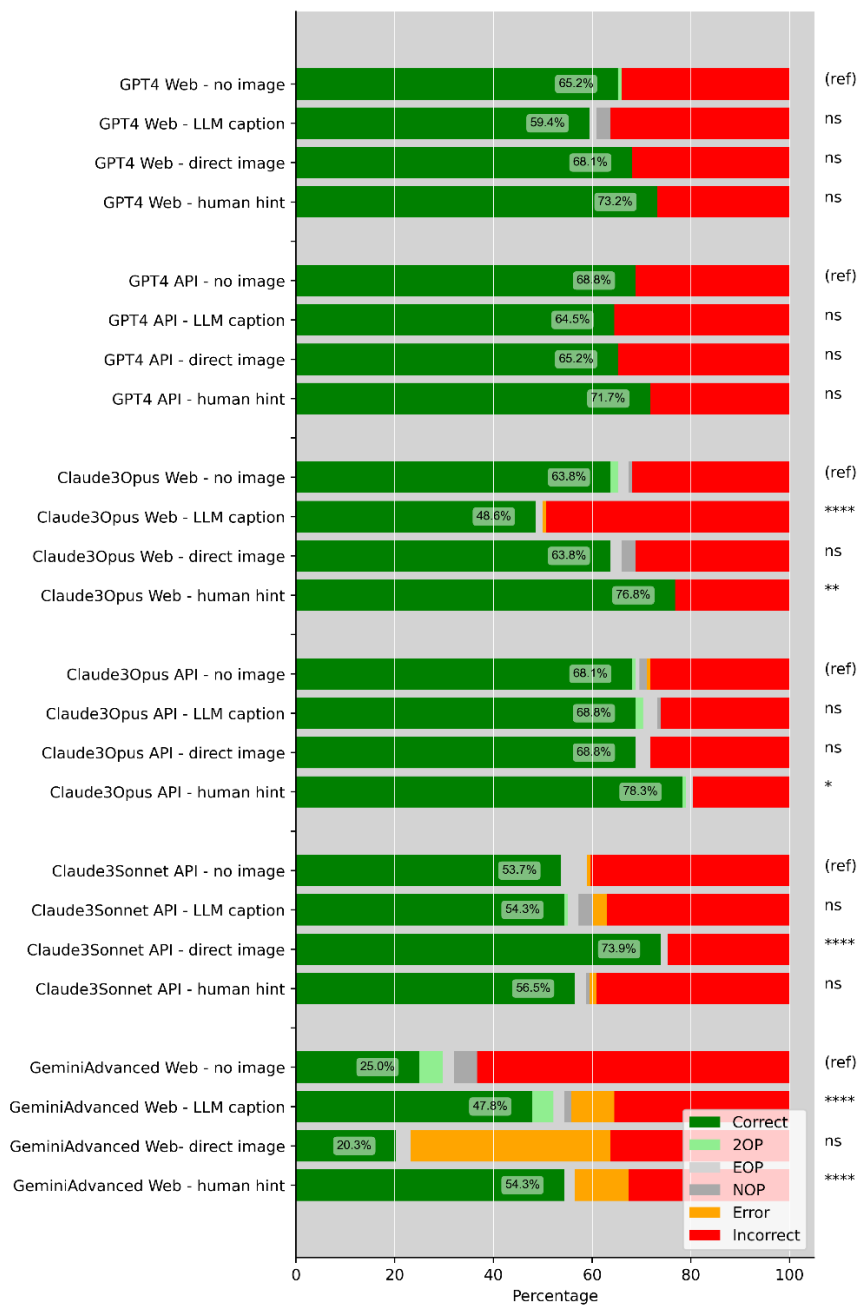

**Figure 5. VLM Performance on Gastroenterology Multiple-Choice Questions with Image Data.**

*Image Description: Performance of three vision language models in two environments in four scenarios. The first bar shows the "no image" scenario where models only have the question stem without any image information, representing the baseline performance. The second bar displays the scenario where captions are generated via VLM, and then the question stem and VLM-generated captions are provided. The third bar shows the scenario of directly providing the image along with the question stem. The fourth bar shows the scenario of providing a short human hint derived from the image along with the question stem, representing the maximum performance when having the ground truth of the image.*

## Historic Performance of GPT Models

LLM performance varied on the 2021, 2022, and 2023 versions of the examination, with GPT-4 scoring 71.59%, 65.38%, and 70.35%, respectively, while GPT-3.5 scoring 44.57%, 46.30%, and 53.49%, respectively (**Figure 6a**). As illustrated in **Figure 6.b**, updating the model's knowledge with more recent training data did not result in performance enhancement. However, each successive model version—from the base GPT to GPT-3.5 (including the GPT-3.5 Turbo), GPT-4, and GPT-4o—demonstrated incremental improvements in performance.

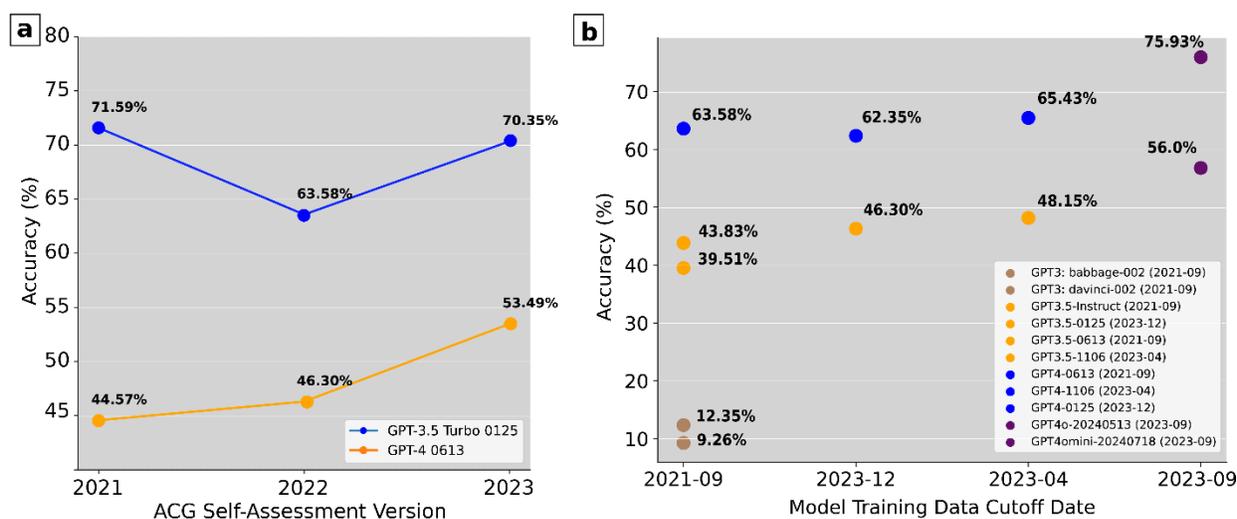

**Figure 6. Historical performance of GPT models on gastroenterology MCQ: (a) three ACG datasets published in 2021, 2022, and 2023 and (b) training data cutoff dates of GPT models.**
*Image Description: The left graph (a) shows the performance of the GPT-3.5 and GPT-4 models on three versions of the ACG-SA dataset. The right graph (b) shows the performance of the GPT models during their evolution, considering the model training data date. We grouped GPT models considering their architecture.*

## Auxiliary Results

### Accuracy Consistency

During our setup experiments, we noted inconsistent model outputs across multiple model runs. To explore the impact of key model settings on model consistency, we measured differences in model accuracy over 3 runs (**Figure 7**). We found that a fixed random seed, a temperature of zero, and simpler prompts lead to the most consistent outputs. Notably, an increase in consistency did not necessarily lead to improved accuracy.

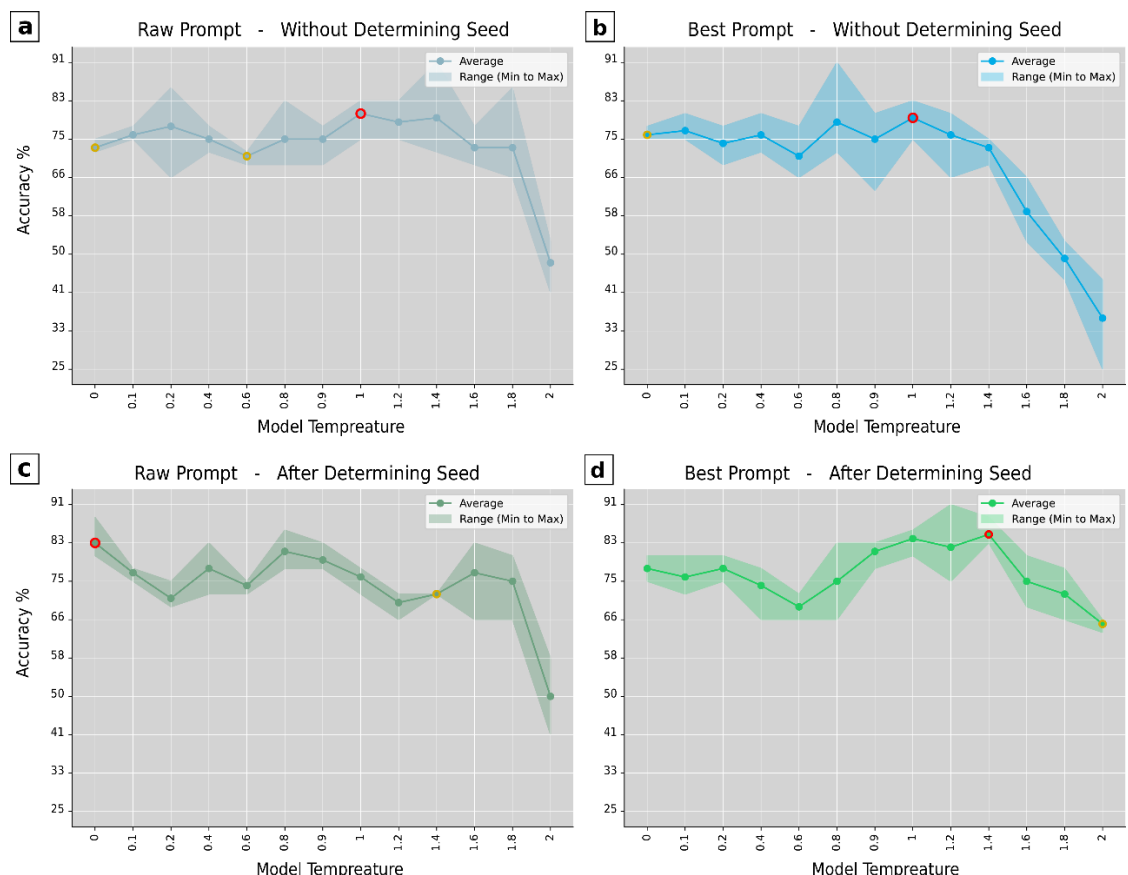

**Figure 7. The impacts of temperature, prompt design, and random seeds on GPT-3.5 consistency are important.**

*Image description: The image shows the number of correct answers in three runs of GPT-3.5 without setting the seed parameter (a and b) and when the seed parameter is determined (c and d). It also depicts the impact of using a simple raw prompt (a and c)* versus *an optimized, engineered prompt (b and d). The shaded areas indicate the minimum and maximum values, and the dots represent the average number of correct answers across the three runs. The temperature with the highest accuracy is highlighted by a red outline, and the temperature with the highest consistency (lowest minimum-maximum range) is highlighted by an orange outline.*

**Estimated cost and run-time**

For a case study, we selected a question consisting of 217 tokens (139 words, 793 characters) characterized by medium difficulty and focused on the medical investigation of liver disease. After incorporating our prompt-engineered command, the total model input increased to 311 tokens (212 words, 1,302 characters). **Supplementary Table S5** details the execution time and cost per token and per execution, along with the minimum required infrastructure for all the setups in Experiment 1.

# Discussion

Our study aimed to assess the performance of LLMs and VLMs in gastroenterology via a reproducible methodology, providing initial insights for future applications of these tools. The capabilities of LLMs have demonstrated remarkable progress, with accuracy rates rising from 12% in base GPT-3 architectures to an impressive 76% in GPT-4. Notably, the Claude 3 Opus, the most advanced model in the Claude 3 family, achieved a performance of 66%, equal to that of GPT-4. However, the best-performing Mistral and Gemini models have not yet reached this level, with accuracy rates ranging between 50% and 55%.

We experiment with a selection of open-source models, including five based on the Llama2, Llama3, and Mixtral architectures. Llama3.1-405b achieved 64% accuracy, reaching GPT-4 performance, followed by Mixtral 8x7b and Llama3-70b with 55% accuracy, and smaller models of Llama3.1-8b with 43% accuracy. While open-source models have shown lower performance than commercial models do, these models can be run locally and offer multiple advantages in the medical field, including data privacy and flexibility for use [28]. However, running these LLMs locally requires infrastructure that typically exceeds the typical capabilities of current healthcare systems.

For example, even a relatively small model with 7 billion parameters requires approximately 28 gigabytes of memory, not including the memory consumed by the operating system [29]. Larger models, with 13 billion or 70 billion parameters, demand approximately 52 GB and 280 GB of memory, respectively. Moreover, achieving acceptable inference times (i.e., the time taken to process a prompt and generate a response) necessitates processing power that is often beyond the reach of many users in the medical field. Utilizing cloud-based services may impose similar limitations to those of commercial models, leaving users with two options: either investing significantly in the necessary infrastructure, which is particularly challenging in the developing world, or opting for smaller models that require less memory.

Quantization offers a potential solution by reducing the model size at the cost of some performance loss. A detailed explanation of this process is beyond the scope of our study (for more information, see Huggingface and Jacob *et al*.) [29], [30]. In essence, quantization involves converting a full-precision model (with 32-bit weights) to a lower-precision version, such as 16-bit (Q16) or 8-bit (Q8), resulting in models that are half or a quarter of the original size. This makes running LLMs on a single local machine feasible and reduces energy consumption. Although the extent of performance loss due to quantization is not yet well established in the medical domain, this approach allows users to maintain full control over their data, thereby safeguarding patient information.

In our study, we evaluated the performance of seven quantized models on a local machine accessible to general users. The Phi3-14b-Q6 model was the top performer, achieving an accuracy of 48.7%, followed by the Mistral-7b-Q8 model. While the performance drop from 43% to 31% was notable for the Llama3-8b-Q8 model, three other quantized models—Mistralv2-7b-Q8, Llama2-13b-Q5, and Gemma2-9b-Q8—maintained results similar to those of their full-size versions. Additionally, fine-tuned versions of Llama2-7b (medicineLLM) and Llama3-8b (OpenBioLLM) did not show comparable or improved performance when run in their 8-bit quantized forms. This suggests that fine-tuning LLMs may not completely mitigate the performance loss associated with quantization.

Previous studies on nonmedical tasks have indicated that performance degradation during 8-bit quantization is minimal [31], [32]. Our findings align with these studies, as three out of four quantized models showed minimal to no drop in performance. Comparing Phi3-3b-Q18 (42%) and Phi3-14b-Q6 (48%), it appears that using larger quantized models may be preferable to running full-precision versions of smaller models. However, these conclusions are based on our limited investigation of performance drop during quantization, which warrants further experimentation. Future studies should consider the impacts of machine configurations, different quantization levels, and performance across different tasks and domains, including multimodal models.

Medical tasks often involve the integration of images and text, making VLMs potentially valuable tools. However, in our experiments, all three VLMs tested—GPT-4-V, Claude3-Opus-V, and GeminiAdvanced-V— failed to demonstrate effective medical image understanding. Their performance did not improve when images were directly analyzed alongside questions, either by generating captions or by using those captions with the questions. Interestingly, the accuracy of the VLMs significantly improved when we provided one-sentence human hints about the images, which revealed some ground truth without directly pointing to the answer. Consistently, a study of GPT-4V performance in medical cases revealed a decrease in accuracy from 80% without text to 58% when images were provided along with the text, emphasizing that either the retrieved information was incorrect or the added complexity outweighed the informational value [33].

Evaluating the accuracy of LLMs via MCQs is only the initial step in their assessment. While MCQs enable quick accuracy evaluations, they have several limitations. First, the real-world application of LLMs is more dynamic and involves complex interactions between physicians and LLMs, as well as between physicians and patients. Second, the role of LLMs extends beyond merely answering clinical scenario questions; they can extract data, assist in documentation, summarize information, and facilitate patient reviews [34]. Third, evaluating LLM responses requires a nuanced approach. Even when an LLM selects the correct option, it may exhibit hallucinations in its response. Conversely, an incorrect answer might still provide valuable cautionary advice or insightful information on differential diagnoses and the next steps.

Beyond the limitations of using MCQs, LLMs should be evaluated using additional metrics beyond mere accuracy. For example, addressing the issue of model hallucination requires methods to measure model certainty [35]. As discussed by Liang *et al*., these metrics include consistency (providing similar answers in repeated runs), calibration (understanding their uncertainty), efficiency (balancing time, cost, energy, and carbon emissions with accuracy), robustness (handling typos and errors in real-world practice), and fairness [6].

Our study revealed challenges with consistency and instances of hallucination in language model outputs. We observed a ±5% variation when repeating the same experiment with GPT-3.5, aligning with our previous findings on model uncertainty in LLM performance [36]. OpenAI has introduced a model setting called "seed" to address such inconsistencies. Additionally, many user guides suggest that setting lower temperatures can result in more deterministic outputs. Our initial results indicate that while setting a seed can lead to more reproducible outputs, the model's behavior with respect to temperature settings remains unpredictable. Furthermore, the optimal setup for achieving the highest accuracy differs from that for achieving the highest consistency. A previous study on USMLE soft skills revealed that GPT-3.5 and GPT-4 changed their answers in a follow-up question in 82% and 0% of cases, respectively [24]. These controversial findings suggest that future investigations are warranted to explore these aspects in greater depth.

An organizational effort is needed to develop an automated, validated, and reproducible pipeline for evaluating LLMs and VLMs in medicine. While our efforts to create such a pipeline represent progress, they also highlight the need for collaboration among multiple parties and governing bodies. The primary reason for using MCQs and focusing on accuracy in this study, as well as in much of the existing research, is the lack of access to a more suitable validation dataset [6]. Ideally, this dataset should reflect the real-world usage of LLMs in medicine and gastroenterology and encompass metrics beyond accuracy.

Several considerations should be taken into account when interpreting the results of this study. Although we integrated a setup experiment into the pipeline, we limited our configuration to OpenAI models and reused the best model settings for other LLMs. A more effective approach would involve creating unique sets of prompts and model settings for each individual LLM. Additionally, the use of quantized models in local environments is a nascent field, with many libraries and modules still in beta development. This is also true for the quantized models accessed through the Hugging Face portal. We have already noted the potential information loss when evaluating LLM responses as simply correct or incorrect. Another limitation of our study is the limited number of experimental runs. Future research would benefit from conducting multiple runs and reporting the average and SD, as the SD can provide insights into model consistency. Finally, our resources constrained the number of LLMs we could test, focusing only on a handful of models that were accessible and well publicized. Consequently, many other models that could offer better performance were not included in our study.

## Conclusion

This research delved into the practical application of LLMs and VLMs in the medical domain, comparing the performance of multiple open-source and quantized models against commercial alternatives. Our study illustrates the importance of optimizing model settings, highlighting concerns such as consistency, hallucination, cost, and speed efficiency. However, while evaluating LLMs via MCQs is straightforward and accessible, it may not fully capture their relevance to clinicians' workflows. Therefore, organizational efforts are crucial to define and develop benchmark datasets and evaluation pipelines to better reflect real-world clinical needs. We plan future refinement of our pipeline to ensure a systematic and reproducible assessment of LLM performance, fostering continuous improvement and adaptation in the field.

**Conflict of interest declaration**

SAASN: Received nonsignificant financial compensation in 2023 as an R&D associate of Aryasp Co.; SA: None; OS: None; ZS: None; TS: None; SR: None; JSS: None; RAS: None; FL: None; JOY: None; JE: None; SB: None; ASh: None; SM: None; NPT: None; GN: None; BEK: None; ASo: None.

**Acknowledgments**

We want to thank the American College of Gastroenterology for kindly providing their valuable question bank, the Hugging Face team for making AI accessible and easy-to-use, and the Bloke account on Hugging Face for providing quantized versions of open-source LLMs.

**Declaration of Generative AI and AI-assisted Technologies**

During the preparation of this work, the author used GPT4 to improve the English style and grammar of the manuscript. After using this tool/service, the authors reviewed and edited the content as needed and take full responsibility for the content of the publication.

**Funding**

No funding was available related to this work.

**Authors' Contribution**

SAASN: Conceptualization, Methodology, Formal Analysis, Investigation, Data Curation, Writing Original Draft, Programming, Project Administration; SA: Methodology, Investigation; OS: Methodology, Investigation; ZS: Investigation; TS: Validation, Investigation, Writing, Review & Editing; SR: Investigation; JSS: Investigation; RAS: Investigation; FL: Investigation; JOY: Investigation; JE: Investigation; SB: Investigation; ASh: Investigation; SM: Validation, Writing Original Draft, Writing Review & Editing; NPT: Validation, Writing, Review & Editing; GN: Methodology, Validation, Writing Review & Editing, Supervision; BEK: Conceptualization, Methodology, Resources, Data Curation, Writing Review & Editing, Project Administration; ASo: Conceptualization, methodology, formal analysis, resources, data curation, writing original draft, writing review & editing, programming, supervision, project administration

**Data availability statement**

The underlying code for this study is available in GitHub repository and can be accessed via this link https://github.com/Sdamirsa/LLM-VLM-in-Gastroenterology. The data that support the findings of this study are available from American College of Gastroenterology (ACG) but restrictions apply to the availability of these data, which were used under license for the current study, and so are not publicly available. Data are however available from the authors upon reasonable request and with permission of ACG. In addition, the ACG self-assessment question and answers are available for members at https://education.gi.org/.

# References

[1]	J. B. Henson, J. R. Glissen Brown, J. P. Lee, A. Patel, and D. A. Leiman, 'Evaluation of the Potential Utility of an Artificial Intelligence Chatbot in Gastroesophageal Reflux Disease Management', *American Journal of Gastroenterology*, vol. 118, no. 12, pp. 2276–2279, Dec. 2023, doi: 10.14309/ajg.0000000000002397.

[2]	S. A. A. Safavi-Naini, Z. Tajabadi, B. El Kurdi, J. A. Abrams, G. N. Nadkarni, and A. Soroush, 'Su1962 GI-COPILOT: AUGMENTING CHATGPT WITH GUIDELINE-BASED KNOWLEDGE', *Gastroenterology*, vol. 166, no. 5, p. S-882, 2024.

[3]	J. Clusmann *et al.*, 'The future landscape of large language models in medicine', *Communications Medicine*, vol. 3, no. 1, Oct. 2023, doi: 10.1038/s43856-023-00370-1.

[4]	L. Tang *et al.*, 'Evaluating large language models on medical evidence summarization', *NPJ Digit Med*, vol. 6, no. 1, Dec. 2023, doi: 10.1038/s41746-023-00896-7.

[5]	S. Ali *et al.*, 'Large Language Models Are Poor Medical Coders — Benchmarking of Medical Code Querying', *NEJM AI*, vol. 1, no. 5, p. AIdbp2300040, Apr. 2024, doi: 10.1056/AIdbp2300040.

[6]	P. Liang *et al.*, 'Holistic Evaluation of Language Models', Nov. 2022, [Online]. Available: http://arxiv.org/abs/2211.09110

[7]	E. Klang, A. Sourosh, and G. N. Nadkarni, 'Evaluating the role of ChatGPT in gastroenterology: a comprehensive systematic review of applications, benefits, and limitations', Jan. 01, 2023, *SAGE Publications Ltd*. doi: 10.1177/17562848231218618.

[8]	K. Singhal *et al.*, 'Towards Expert-Level Medical Question Answering with Large Language Models', May 2023, [Online]. Available: http://arxiv.org/abs/2305.09617

[9]	W. Li, L. Li, T. Xiang, X. Liu, W. Deng, and N. Garcia, 'Can multiple-choice questions really be useful in detecting the abilities of LLMs?', Mar. 2024, [Online]. Available: http://arxiv.org/abs/2403.17752

[10] A. Gilson *et al.*, 'How Does ChatGPT Perform on the United States Medical Licensing Examination? The Implications of Large Language Models for Medical Education and Knowledge Assessment', *JMIR Med Educ*, vol. 9, 2023, doi: 10.2196/45312.

[11] T. H. Kung *et al.*, 'Performance of ChatGPT on USMLE: Potential for AI-assisted medical education using large language models', *PLOS Digital Health*, vol. 2, no. 2, p. e0000198, Feb. 2023, doi: 10.1371/journal.pdig.0000198.

[12] M. Moshirfar, A. W. Altaf, I. M. Stoakes, J. J. Tuttle, and P. C. Hoopes, 'Artificial Intelligence in Ophthalmology: A Comparative Analysis of GPT-3.5, GPT-4, and Human Expertise in Answering StatPearls Questions', *Cureus*, Jun. 2023, doi: 10.7759/cureus.40822.

[13] A. Mihalache, M. M. Popovic, and R. H. Muni, 'Performance of an Artificial Intelligence Chatbot in Ophthalmic Knowledge Assessment', *JAMA Ophthalmol*, vol. 141, no. 6, pp. 589–597, Jun. 2023, doi: 10.1001/jamaophthalmol.2023.1144.

[14] L. Z. Cai *et al.*, 'Performance of Generative Large Language Models on Ophthalmology Board–Style Questions', *Am J Ophthalmol*, vol. 254, pp. 141–149, Oct. 2023, doi: 10.1016/j.ajo.2023.05.024.

[15] K. Suchman, S. Garg, and A. J. Trindade, 'Chat Generative Pretrained Transformer Fails the Multiple-Choice American College of Gastroenterology Self-Assessment Test', *American Journal of Gastroenterology*, vol. 118, no. 12, pp. 2280–2282, Dec. 2023, doi: 10.14309/ajg.0000000000002320.

[16] R. Noda, Y. Izaki, F. Kitano, J. Komatsu, D. Ichikawa, and Y. Shibagaki, 'Title Performance of ChatGPT and Bard in Self-Assessment Questions for Nephrology Board Renewal Author', doi: 10.1101/2023.06.06.23291070.

[17] I. Skalidis *et al.*, 'ChatGPT takes on the European Exam in Core Cardiology: an artificial intelligence success story?', *European Heart Journal - Digital Health*, vol. 4, no. 3, pp. 279–281, May 2023, doi: 10.1093/ehjdh/ztad029.

[18] L. Passby, N. Jenko, and A. Wernham, 'Performance of ChatGPT on Specialty Certificate Examination in Dermatology multiple-choice questions', *Clin Exp Dermatol*, Jun. 2023, doi: 10.1093/ced/llad197.

[19] P. Humar, M. Asaad, F. B. Bengur, and V. Nguyen, 'ChatGPT Is Equivalent to First-Year Plastic Surgery Residents: Evaluation of ChatGPT on the Plastic Surgery In-Service Examination', *Aesthet Surg J*, vol. 43, no. 12, pp. NP1085–NP1089, Dec. 2023, doi: 10.1093/asj/sjad130.

[20] C. C. Hoch *et al.*, 'ChatGPT's quiz skills in different otolaryngology subspecialties: an analysis of 2576 single-choice and multiple-choice board certification preparation questions', *European Archives of Oto-Rhino-Laryngology*, vol. 280, no. 9, pp. 4271–4278, Sep. 2023, doi: 10.1007/s00405-023-08051-4.

[21] Z. C. Lum, 'Can Artificial Intelligence Pass the American Board of Orthopaedic Surgery Examination? Orthopaedic Residents Versus ChatGPT', *Clin Orthop Relat Res*, vol. 481, no. 8, pp. 1623–1630, Aug. 2023, doi: 10.1097/CORR.0000000000002704.

[22] R. Ali *et al.*, 'Performance of ChatGPT, GPT-4, and Google Bard on a Neurosurgery Oral Boards Preparation Question Bank', *Neurosurgery*, vol. 93, no. 5, pp. 1090–1098, Nov. 2023, doi: 10.1227/neu.0000000000002551.

[23] H. Khorshidi *et al.*, 'Application of ChatGPT in multilingual medical education: How does ChatGPT fare in 2023's Iranian residency entrance examination', *Inform Med Unlocked*, vol. 41, p. 101314, 2023, doi: https://doi.org/10.1016/j.imu.2023.101314.

[24] D. Brin *et al.*, 'Comparing ChatGPT and GPT-4 performance in USMLE soft skill assessments', *Sci Rep*, vol. 13, no. 1, p. 16492, Oct. 2023, doi: 10.1038/s41598-023-43436-9.

[25] Anthropic Official Website, 'Consumer Terms of Service [Date Published: February 3, 2024; Date Accessed: April 10, 2024]'. Accessed: May 10, 2024. [Online]. Available: https://www.anthropic.com/legal/consumer-terms

[26]  Poe Official Website, 'Poe Privacy Center [Date Published: NA; Date Accessed: April 10, 2024]'. Accessed: May 10, 2024. [Online]. Available: https://poe.com/privacy_center

[27]  OpenAI Official Website, 'Data Controls FAQ [Last update: April 2024; Date Accessed: April 10, 2024]'. Accessed: May 10, 2024. [Online]. Available: https://help.openai.com/en/articles/7730893-data-controls-faq

[28]  Y. Zheng, W. Gan, Z. Chen, Z. Qi, Q. Liang, and P. S. Yu, 'Large Language Models for Medicine: A Survey', *arXiv preprint arXiv:2405.13055*, 2024.

[29]  Hugging Face, 'Qunatization'. Accessed: May 13, 2024. [Online]. Available: https://huggingface.co/docs/optimum/en/concept_guides/quantization

[30]  B. Jacob *et al.*, 'Quantization and Training of Neural Networks for Efficient Integer-Arithmetic-Only Inference', *2018 IEEE/CVF Conference on Computer Vision and Pattern Recognition*, pp. 2704–2713, 2017, [Online]. Available: https://api.semanticscholar.org/CorpusID:39867659

[31]  S. Li *et al.*, 'Evaluating Quantized Large Language Models', *ArXiv*, vol. abs/2402.18158, 2024, [Online]. Available: https://api.semanticscholar.org/CorpusID:268041618

[32]  Z. Gong, J. Liu, J. Wang, X. Cai, D. Zhao, and R. Yan, 'What Makes Quantization for Large Language Model Hard? An Empirical Study from the Lens of Perturbation', *ArXiv*, vol. abs/2403.06408, 2024, [Online]. Available: https://api.semanticscholar.org/CorpusID:268357772

[33]  T. A. Buckley, J. A. Diao, A. Rodman, and A. K. Manrai, 'Accuracy of a Vision-Language Model on Challenging Medical Cases', *ArXiv*, vol. abs/2311.05591, 2023, [Online]. Available: https://api.semanticscholar.org/CorpusID:265067031

[34]  O. Shahab, B. El Kurdi, A. Shaukat, G. Nadkarni, and A. Soroush, 'Large language models: a primer and gastroenterology applications', Jan. 01, 2024, *SAGE Publications Ltd*. doi: 10.1177/17562848241227031.

[35]  S. Lin, J. Hilton, and O. Evans, 'Teaching Models to Express Their Uncertainty in Words', May 2022, [Online]. Available: http://arxiv.org/abs/2205.14334

[36]  T. Savage *et al.*, 'Large Language Model Uncertainty Measurement and Calibration for Medical Diagnosis and Treatment', *medRxiv*, p. 2024.06.06.24308399, Jan. 2024, doi: 10.1101/2024.06.06.24308399.

[37]  American Board of Internal Medicine (ABIM) Official Website, 'GASTROENTEROLOGY CERTIFICATION EXAM CONTENT'. Accessed: May 14, 2024. [Online]. Available: https://www.abim.org/certification/exam-information/gastroenterology/exam-content

[38]  T. Dettmers, A. Pagnoni, A. Holtzman, and L. Zettlemoyer, 'QLoRA: Efficient Finetuning of Quantized LLMs', in *Advances in Neural Information Processing Systems*, A. Oh, T. Naumann, A. Globerson, K. Saenko, M. Hardt, and S. Levine, Eds., Curran Associates, Inc., 2023, pp. 10088–10115. [Online]. Available: https://proceedings.neurips.cc/paper_files/paper/2023/file/1feb87871436031bdc0f2beaa62a049b-Paper-Conference.pdf

[39]  B. Xiao, 'Some basic knowledge of LLM: Parameters and Memory Estimation', Medium. Accessed: May 14, 2024. [Online]. Available: https://medium.com/@baicenxiao/some-basic-knowledge-of-llm-parameters-and-memory-estimation-b25c713c3bd8

[40]  ggml authors (Github Repository: ggerganov/ggml), 'GGUF'. Accessed: May 15, 2024. [Online]. Available: https://github.com/ggerganov/ggml/blob/master/docs/gguf.md

[41]  M. Kowalczyk, 'A step-by-step guide to prompt engineering: Best practices, challenges, and examples', Lakera.ai. Accessed: Mar. 15, 2024. [Online]. Available: https://www.lakera.ai/blog/prompt-engineering-guide

[42]  B. Meskó, 'Prompt Engineering as an Important Emerging Skill for Medical Professionals: Tutorial', *J Med Internet Res*, vol. 25, no. 1, Jan. 2023, doi: 10.2196/50638.

[43]   Y. Zhou *et al.*, 'Large Language Models Are Human-Level Prompt Engineers', Nov. 2022, [Online]. Available: http://arxiv.org/abs/2211.01910

[44]   N. Zaidi, 'Modified Bloom's Taxonomy for Evaluating Multiple Choice Questions', Baylor College of Medicine. Accessed: May 20, 2024. [Online]. Available: https://www.bcm.edu/sites/default/files/2019/04/principles-and-guidelines-for-assessments-6.15.15.pdf

by Seyed Amir Ahmad Safavi-Naini, Shuhaib Ali, Omer Shahab, Zahra Shahhoseini, Thomas Savage , Sara Rafiee, Jamil S. Samaan, Reem Al Shabeeb, Farah Ladak, Jamie O. Yang, Juan Echavarria, Sumbal Babar, Aasma Shaukat, Samuel Margolis, Nicholas P. Tatonetti, Girish Nadkarni, Bara El Kurdi4, Ali Soroush

Corresponding to Ali Soroush (Ali.Soroush@mountsinai.org) and Bara El Kurdi (baraabayern@gmail.com).

## List of Supplementary Sections

# Supplementary Section 1: Exam and Question Context

## 1.1 Gastroenterology Board Exam Context

The Gastroenterology Board Exam is a rigorous and essential assessment that gastroenterologists must pass to become board-certified in their specialty. Administered by the American Board of Internal Medicine, this exam tests a candidate's knowledge, practical skills, and critical thinking abilities in gastroenterology. The exam ensures that practitioners meet the high standards required to provide quality care in this complex field. Test takers undergo a rigorous educational pathway before taking the exam, including four years of undergraduate study, four years of medical education, three years of internal medicine residency, and three years of gastroenterology fellowship. By passing this exam, these physicians achieve board certification and affirm their expertise in providing high-quality care in gastroenterology [37].

## 1.2 ACG Question Bank Characteristics

The American College of Gastroenterology (ACG) developed the Board Prep/Self-Assessment Test (SA), which was developed meticulously by an ACG committee with contributions from most postgraduate course faculty members. This comprehensive test consists of approximately 300 questions annually. These questions, presented in single best answer and true/false formats and often supplemented with photos and illustrations, come with in-depth answers and referenced annotations. It offers an essential review for both graduating fellows preparing for their GI Boards and seasoned practitioners seeking a comprehensive clinical update.

**Supplementary Table S1.** Lists of glossaries for the model environment and parameters, along with the exact model name and date of use.

| Term | Description |
|---|---|
| Large Language Model (LLM) | A type of advanced artificial intelligence model designed to understand, generate, and manipulate human language. LLMs are trained on vast amounts of textual data, allowing them to perform a wide range of language-based tasks such as text completion, translation, summarization, and question answering. These models leverage deep learning techniques, particularly transformer architectures, to analyze and produce language at a scale and complexity that closely mimics human linguistic capabilities. |
| Vision Language Model (VLM) | A type of artificial intelligence model that integrates and processes both image and textual data to perform tasks that require understanding and reasoning across these two modalities. |
| Model Parameters (e.g. 7B) | This term refers to the total count of trainable weights and biases in a neural network model. Each parameter represents a component of the model that can learn from data during the training process. For example, the notation "7B" indicates that the model has 7 billion parameters [38]. |
| Quantization and Quantized Models | This process involves reducing the precision of numbers that represent data in LLMs. Typically, models use 32-bit precision, but quantization reduces this to 16-bit, 8-bit, or 4-bit. The main goal is to cut down on the amount of memory and power the models use, making them more efficient and suitable for devices with limited resources [38], [39]. |
| Required Memory for Running LLMs | The amount of memory needed to run a LLM varies based on several factors. As a simplified rule, for a full precision 32-bit model, multiply the number of model parameters by 4 to estimate the required memory in bytes. For a quantized model, such as an 8-bit version, the memory requirement can be reduced. Since 8-bit precision uses one-quarter of the memory that 32-bit precision does, you can divide the memory estimate for a full precision model by 4 to calculate the memory needed for an 8-bit quantized model [39]. |
| GGUF Model Format | GGUF, or GPT-Generated Unified Format, is a file format specifically crafted for storing inference data of language models. GGUF streamlines the distribution of models by integrating all necessary components into a single file, thereby obviating the need for additional external files. It enhances flexibility by allowing the integration of new features and information into existing models without affecting their compatibility. A significant benefit of GGUF is its capability to run LLM locally, which is particularly advantageous when these models exceed the memory limitations of GPUs. As a comprehensive, user-friendly, and CPU-compatible format, GGUF proves especially valuable for local operation of LLMs, facilitating ease and efficiency in handling complex language processing tasks [40]. |
| Environment | The platform user can utilize to use LLMs and talk to chatbots. |
| Environment: Web | This environment refers to a web-based platform where users can interact with LLMs and chatbots directly through their internet browsers. Users typically provide inputs (prompts) to the model without the capability to control detailed model parameters. |
| Environment: API | In this scenario, the LLM or chatbot is accessed through an Application Programming Interface (API). More capabilities and control over model parameters are achievable when using APIs. |
| Environment: Local | A local environment means running the LLM or chatbot directly on a user's computer or on a private server. This setup requires installing and managing the model on local hardware. It offers the highest level of control over data privacy and system configuration but demands more technical knowledge and resources to maintain. |
| LLM Setting | LLM settings, sometimes referred to as parameters, hyperparameters, or configurations, consist of changeable variables that govern the behavior of a model at the time of use and are not learned from the training data. |
| LLM Setting: temperature | Temperature is a parameter used to control the randomness of predictions by scaling the logits.<br>Low temperature (close to 0): The model's outputs become more deterministic and repetitive, with higher probability being assigned to the most likely words. This setting makes the model more predictable and often more coherent, but less diverse in its responses.<br>High temperature: The model outputs become more random and diverse. High temperatures can make the text more varied and creative but also more likely to produce nonsensical or irrelevant outputs. |
| LLM Setting: max_token | Max token refers to the maximum number of tokens (words or subwords) the model can generate in a single input-plus-output. It helps control the length of generated text. |
| LLM Setting: seed | Seed refers to a starting point or initial input given to the model, which influences the subsequent text generation. It can be a specific word, phrase, or context provided by the user. |
| LLM Setting: user prompt (prompt) | User prompt, often referred to simply as prompt, is the input provided by the user to the model to generate text. It can be a question, statement, or context intended to guide the model's response. |
| LLM Setting: system prompt | System prompt refers to a predefined or automated input provided to the model by the system or application to initiate text generation and maintain consistency in the conversation. |

# Supplementary Section 2: Full detail of Experiment Zero

**Prompt:** The prompt is the instruction given to the LLM to elicit answers. Initially, we used a basic prompt (e.g., "select the correct option from the provided options") and subsequently tested 11 different prompt engineering techniques individually. We identified the most effective techniques and developed six candidates for the optimal prompt via a fully automated pipeline with the GPT-4 API (see **Supplementary Table S3**).

Our approach involved sending individual prompts in separate requests in the API and web environment; thus, we did not test system prompts, as all the context is provided in the message. System prompts are designed to provide the model with the necessary context for specific tasks, optimizing the interaction of conversational LLMs.

**Maximum Token Count (max_token)**: This setting determines the allowable length of the input and output total for the LLMs. We tested maximum token limits of 512, 1024, and 512 plus (i.e., adding 512 tokens to the model's input, allowing a fair response space across varying question stem lengths). Importantly, responses may sometimes be truncated at these limits, potentially leaving them incomplete, which is considered a technical limitation (**Supplementary Table S4**).

**Temperature**: The temperature controls the randomness of the model's responses. A lower temperature (close to 0) results in more deterministic and repetitive outputs, with a higher likelihood assigned to the most likely words. Conversely, a higher temperature leads to more random and diverse outputs, which can enhance creativity but also increase the probability of nonsensical or irrelevant responses (**Supplementary Table S5**).

## 2.1 Finding the Best Setting: Function, Prompt, Max-Token, and Temperature

The aim of this experiment was to find the best set of model settings and set up future experiments. This experiment was conducted via a random sample of 60 questions from 2021 and 2023. We used GPT3 (gpt-3.5-turbo-0125 in March 2024) to find the best setup and replicate this setup in future experiments. The step-by-step model settings include:

- **Function**: Investigating the impact of obtaining a structured output via an OpenAI function call or a Lang chain output parser versus obtaining an unstructured answer to the question. The results and descriptions are available in Supplementary Table S4.

- **Prompt:** To find the best prompt utilizing prompt engineering techniques, we followed a series of replicable steps, which are fully described in **Supplementary Table S5**.
    - **Raw prompting:** The prompt was 'Answer the following question and select one option from the provided options.'
    - **Prompt engineering techniques, one at a time:** Using the resources for prompt engineering techniques, we first tested the impact of each prompt engineering approach from the literature and then selected the techniques with improvement in performance [41], [42]. Compared with raw accuracy, effective prompt engineering techniques capable of increasing performance were selected. To exclude randomness and find effective prompt engineering techniques, the experiment was run three times for each prompt. Effective prompts were selected when the average performance of each prompt surpassed the upper bound of the 95% confidence interval (95% CI) of the raw prompts in three runs.
    - **Best prompt, utilizing effective prompt engineering techniques:** Inspired by Zhou *et al*. [43], we provided the selected techniques to GPT4 with prompts. We also created three other versions, generated via this prompt to GPT4 "Act as an expert data scientist in the field of medicine with deep knowledge of prompt engineering and LLMs. Improving this prompt and providing the most concise and clear prompt for a large language model." Finally, the best prompt was selected on the basis of the average of three runs for each prompt scenario, including a raw prompt, 11 single prompt engineering techniques, and six best prompt candidates (three runs consisting of two with 512 max_token and one with 1024). The details of the prompt engineering approaches, their definitions, and their impacts on performance are presented in **Supplementary Table S5**.
- **Model temperature:** In LLMs, temperature controls prediction randomness. Lower temperatures yield more predictable text, whereas higher temperatures increase diversity and creativity, potentially at the cost of relevance. The default temperature of 0 and max_token of 512 were utilized in the previous experiments. We tested the performance of temperatures of 0, 0.1, 0.4, 0.6, 0.9, and 1 on the raw prompt and best prompt. The performance is reported in **Supplementary Table S6.**

- **Model max_token:** A token is a basic unit of text, such as a word or punctuation mark, used by LLMs to understand and generate text. We tested the max_token of 512, and as our initial experiment revealed the better performance of max_token 1024 for prompts that need space for generation, we added another solution, which is based on the length of the input text. This max_token points to the total number of input tokens (prompt, question, and options) and output tokens (generated tokens), so in our third scenario, we first counted the number of input tokens and set the max_token of 512+input tokens. Consequently, this approach can provide a fair area for model output, regardless of the question-option length. The performances in these three scenarios are listed in **Supplementary Table S7.**
- **Consistency:** We ran the raw prompt and each prompt engineering technique 3 times via GPT 3.5. In addition to finding effective prompt engineering techniques (when the average performance of a prompt exceeds the 95% CI), we demonstrated consistency among prompt strategies.

## 2.2 Structured Output

We tested the two options of using the Langchain library and the official structured output ability provided by OpenAI and Claude3. This feature was first announced by OpenAI, and Calude3 recently added this feature to their API capability. In addition, the Langchain can force an LLM to provide a response in the desired format so that the parser can parse this response to provide a structured output (JSON or dictionary). Behind the scene of the Langchain, Claude3, or OpenAI function calling (i.e., the feature used to provide structured output), injects some instructions into the message, parses the output, and retrieves parameters from the generated message.

For other models, including all outputs generated in the Web environment, Poe environment, and Local environment, as well as old GPT models (instruct-GPT, babbage, and DaVinci), the output was unstructured. The method for extracting the selected option from the raw output can be found in the "Evaluation of accuracy" section.

## 2.3 Structured and Unstructured Prompt

The full prompt was structured as a command prompt, one empty line, one question, one empty line, and one option. In cases of generating structured output (GPT4-API, GPT3.5-API, Claude3Haiku-API, Claude3Sonnet-API, and Claude3Opus-API), the full prompt was provided along with a schema for function call. The function call is the OpenAI and Claude3 tool, which can generate structure outputs via a schema consisting of a schema description, parameters for output, and parameter definitions. The scheme description in our experiment was "You can generate your unstructured output and selected option." The LLM can output two parameters, "unstructured output" and "selected option", with the corresponding parameter definitions of "Your unstructured output for the answer, if needed." and "The selected option from the list of provided choices. (Example: 'F')". The options were limited to one of the options "A", "B", "C", "D", or "E".

**Supplementary Table S2:** Investigation of the performance of structured output functions, including OpenAI function calling and Lang chain structured output, compared with that of unstructured outputs (named Open Call).

|  | Experiment Setting | Prompt | GPT3-API Accuracy % (correct answers-of-all questions) | GPT4-API Accuracy % (correct answers-of-all questions) |
|---|---|---|---|---|
| OpenAI Open Call (unstructured) | We used OpenAI to provide unstructured answer and then run function calling to find the correct answer in the answer text (temperature=0). | "Answer the following question and select one option from the provided options." | 40.00% (24-of-60) | 55.00% (33-of-60) |
| OpenAI Function Call (structured output) | We used OpenAI Function Call (temperature=0, max_token =512) | "Answer the following question and select one option from the provided options." | 43.33% (26-of-60) | 68.33% (41-of-60) |
| Lang Chain (structured output) | We used Langchain structure output (create_structured_output_runnable) (temperature=0, max_token =512) | "Answer the following question and select one option from the provided options." | 40.00% (24-of-60) | 66.67% (40-of-60) |

**Supplementary Table S3:** Description of each prompt engineering technique, performance, and final prompt set. The performance was shown on a sample of questions from ACG self-assessment data from 2021 (30 questions) and 2023 (30 questions) (1, 2). [Experiment Setting: 3 runs with max_token= 512 in two runs and 1024 in one run. Temperature = 0. Model= gpt-turbo-3.5-0125.]

|  | Prompt | GPT3.5 Accuracy | Performance Change | Effective[a] |
|---|---|---|---|---|
| **Prompt: One prompt engineering technique at a time** | | | | |
|  |  | Average ± SD [95 CI] [Range: min, max] | % vs raw [Mean difference 95CI] | Yes/No |
| **raw** | "Answer the following question and select one option from the provided options." | **43.89±0.96 [95CI: [42.8 44.98], 2] [Range: 43.33, 45.0]** | REF | REF |
| **Direct Questioning:** Simply inputting the MCQ as is to the AI and asking it to choose the best option. This is the most straightforward approach. | "Could you choose the best option from the provided options?" | 48.33±1.67 **95CI**: [46.44 50.22], 2] [**Range**: 46.67, 50.0] | 4.45; Mean Diff CI: [2.78 6.11]; p value: 0.2500 | Yes |
| **Option Analysis:** Separately inputting each option along with the question to the AI, asking for a validation of each choice, and then selecting the one with the most convincing justification. | "Answer the following question and select one option from the provided options. Analyze each option and explain its validity. Which option is the most accurate and why?" | 43.89±0.96 **95CI**: [42.8 44.98], 2] [**Range**: 43.33, 45.0] | 0.0; Mean Diff CI: [-1.11 1.11]; p value: 1.0000 | No |
| **Chain of Thought:** Asking the AI to elaborate on its thought process for each option, leading to a more informed and transparent decision-making process. | "Answer the following question and select one option from the provided options. Please go through each option and detail your reasoning process behind evaluating each one to determine the correct answer." | 48.89±3.47 **95CI**: [44.96 52.82], 2] [**Range**: 45.0, 51.67] | 5.0; Mean Diff CI: [1.11 7.78]; p value: 0.2500 | Yes |
| **Answer and Justify:** Requesting the AI not only to pick an option but also to provide a rationale for its choice, enhancing the reliability of the response. | "Answer the following question and select one option from the provided options. Please select the correct answer and provide a justification for your choice." | 48.33±1.67 **95CI**: [46.44 50.22], 2] [**Range**: 46.67, 50.0] | 4.45; Mean Diff CI: [2.78 6.11]; p value: 0.2500 | Yes |

|  | Prompt | GPT3.5 Accuracy | Performance Change | Effective[a] |
|---|---|---|---|---|
| **Elimination Method:** Instructing the AI to eliminate clearly wrong options first, narrowing down the choices before making a final decision. | "Answer the following question and select one option from the provided options. Start by eliminating the options that are clearly incorrect. After elimination, which option remains as the correct answer? " | 43.34±2.89 **95CI**: [40.07 46.61], 2] [**Range**: 41.67, 46.67] | -0.55; Mean Diff CI: [-2.77 2.78]; p value: 1.0000 | No |
| **Comparative Analysis:** Asking the AI to compare options against each other to identify which one best fits the question's requirements. | "Answer the following question and select one option from the provided options. Compare the following options to identify the most appropriate answer. How does each one measure against the others in terms of accuracy?" | 42.22±0.96 [**95CI**: [41.13 43.31], 2] [**Range**: 41.67, 43.33] | 1.66; Mean Diff CI: [-2.78 - 0.55]; p value: 0.2500 | No |
| **Contextual Embedding:** Providing additional information or context related to the question before asking the AI to make a choice, especially useful if the question is about a specific domain or topic. | "Answer the following question and select one option from the provided options. This multiple-choice question is designed to simulate a gastroenterology board examination, consisting of complex queries aimed at assessing the depth of knowledge of the examinees. Your response should reflect an understanding of advanced gastroenterology concepts." | 46.66±2.89 **95CI**: [43.39 49.93], 2] [**Range**: 43.33, 48.33] | 2.78; Mean Diff CI: [-0.56 5. ]; p value: 0.1797 | Yes |
| **Confidence Scoring:** Asking the AI to rate its confidence in each option being correct, then selecting the one with the highest score. | "Rate your confidence level in the correctness of each of these options from 1 to 10, where 10 is most confident. Which option scores the highest? " | 45.56±1.93 **95CI**: [43.38 47.74], 2] [**Range**: 43.33, 46.67] | 1.67; Mean Diff CI: [-0.56 3.34] p value: 0.5000 | Yes |
| **Expert Mimicry:** Framing the prompt as if an expert in the relevant field is answering the question, thereby leveraging the model's ability to emulate specialist knowledge. | "Imagine you are a seasoned gastroenterologist facing this board exam question. As an expert in the field, which answer would you choose and why?" | 45.56±0.96 **95CI**: [44.47 46.65], 2] [**Range**: 45.0, 46.67] | 1.67; Mean Diff CI: [0.56 2.78]; p value: 0.2500 | Yes |

|  | Prompt | GPT3.5 Accuracy | Performance Change | Effective [a] |
|---|---|---|---|---|
| **Consensus Technique:** Inputting the question and options multiple times, possibly with slight variations, and choosing the most consistently selected option. | "I will present a gastroenterology board exam question several times with slight variations to evaluate consistent selection. What is your choice? Now, consider the question again with the same options. Has your choice changed or remained consistent?" | 40.0±2.89 **95CI**: [36.73 43.27], 2] [**Range**: 38.33, 43.33] | -3.89; Mean Diff CI: [-6.11 - 0.56]; p value: 0.1797 | No |
| **Give Model Time to Think:** Asking the LLM to answer the question, while providing the chance to trigger more reflection and thinking. | "Take a moment to think carefully about this gastroenterology board exam question. Reflect on your medical knowledge and expertise before determining the most accurate answer. Then, answer the following question and select one option from the provided options." | 0.0 [Mean Diff CI: [-2.22 2.78]] [p value: 1.0000] | 0; Mean Diff CI: [-4.11 8.55]; p value=0.271 | No |
| **Prompt: Generate six 'best prompt' candidates [b]** |||||
| 'best prompt' candidate 1 | "Consider this complex gastroenterology board exam question. Examine each of the options provided closely. Delve into a detailed reasoning process for each option, considering the principles and knowledge that apply to gastroenterology. Reflect on how each option aligns or conflicts with these principles to determine the most accurate answer. Share your thought process step by step to arrive at the correct choice." | 38.33±3.34 95CI: [34.55 42.11], 2] [Range: 35.0, 41.67] | -5.55; Mean Diff CI: [-8.89 - 2.22]; p value: 0.2500 | No |
| 'best prompt' candidate 2 | "Here's a challenging question from a gastroenterology board examination. The options are provided below. After carefully considering the question and the options, select the one you believe is correct. Importantly, provide a comprehensive justification for your selection. Your justification should highlight your understanding of gastroenterology concepts and explain why your chosen option stands out as the most accurate." | 48.33±0.0 95CI: 48.33, 48.33] [Range: 48.33, 48.33] | 4.44; Mean Diff CI: [3.33 5.]; p value: 0.2500 | Yes |

|  | Prompt | GPT3.5 Accuracy | Performance Change | Effective[a] |
|---|---|---|---|---|
| 'best prompt' candidate 3 | "This multiple-choice question is part of a simulated gastroenterology board examination, aimed at testing deep knowledge in the field. The options available are provided below. Directly choose the best option based on your expertise in gastroenterology. Consider the complexities and nuances of the field that are relevant to the question as you make your selection." | 43.89±0.96 95CI: [42.8 44.98], 2] [Range: 43.33, 45.0] | 0.0; Mean Diff CI: [-1.11 1.11]; p value: 1.0000 | No |
| 'best prompt' candidate 4 **[selected best prompt]** | "Imagine you are a seasoned gastroenterologist approaching this complex question from the gastroenterology board exam. Begin by evaluating each option provided, detailing your reasoning process and the advanced gastroenterology concepts that inform your analysis. Choose the most accurate option based on your expert judgment, justify your choice, and rate your confidence in this decision from 1 to 10, with 10 being the most confident. Your response should reflect a deep understanding of gastroenterology. " | 48.89±3.47 95CI: [44.96 52.82], 2] [Range: 45.0, 51.67] | 5.0; Mean Diff CI: [1.11 7.78]; p value: 0.1797 | Yes |
| 'best prompt' candidate 5 | "Analyze a gastroenterology board exam question by reviewing each option, applying advanced concepts for informed reasoning. Select and justify the best answer using expert insight, and rate your confidence (1-10). Your response should insightfully address the question, demonstrating comprehensive gastroenterology expertise." | 47.78±1.92 95CI: [45.61 49.95], 2] [Range: 46.67, 50.0] | 3.89; Mean Diff CI: [2.23 6.11]; p value: 0.2500 | Yes |

|  | Prompt | GPT3.5 Accuracy | Performance Change | Effective[a] |
|---|---|---|---|---|
| 'best prompt' candidate 6 | "Approach a complex gastroenterology board exam question by systematically evaluating each answer choice through the lens of advanced gastroenterology principles. Utilize your expertise to identify the most accurate option, provide a rationale for your selection, and assign a confidence rating from 1 to 10. Ensure your response directly resolves the question, demonstrating an extensive understanding of the field." | 48.89±2.55 95CI: [46.51.78], 2] [Range: 46.67, 51.67] | 5.0; Mean Diff CI: [2.78 7.78]; p value: 0.2500 | Yes |

**Footnote:** SD, standard deviation; 95CI, 95% confidence interval; [a], the effective prompts were selected by comparing the average performance of the prompt engineering technique and upper bound of 95CI; [b], The 6 candidates were created using GPT4-API, temperature=0, max_token=512, and prompt= "I seek to identify the optimal prompt for accurately answering multiple-choice questions, specifically those pertaining to the gastroenterology board exam. Previously, various prompt engineering techniques have been evaluated. The methods that demonstrated performance improvement include chain of thought, answer and justification, direct question answering, contextual embedding, expert mimicry, and confidence scoring. Attached, you can find the exact prompt for each technique. Please provide me with three variations of the best prompt to enhance the accuracy of responses to these complex questions, utilizing the aforementioned prompt engineering techniques." The Wilcoxon signed ranked test was used to compare performance in three runs, with the raw prompt used as the reference scenario. For calculation of the mean difference CI, bootstrapping was used with 10000 bootstraps and an alpha of 0.05 (corresponding to the 95% confidence interval).

**Supplementary Table S4:** The experimentation of max_token, which is the total number of maximum tokens (i.e., the smallest word sections that LLMs use to transform words into mathematical representations). By 512 plus, we mean that we will count the input tokens (question + options) and add this to 512, which means that the model can generate 512 tokens in the output. This method is fairer, as longer and shorter questions have the same output space, and the performance is comparable or better than setting max_token to 512 or 1024. [Experiment Setting: 1 run. Temperature = 0. Model= gpt-turbo-3.5-0125]

|  | GPT3.5 - API Max_token = 512 | GPT3.5 - API Max_token = 1024 | GPT3.5 - API Max_token = 512plus |
|---|---|---|---|
|  | Accuracy % (correct answers-of-all questions) | Accuracy % (correct answers-of-all questions) | Accuracy % (correct answers-of-all questions) |
| Raw | 43.33% (26-of-60) | 41.67% (25-of-60) | 45.00% (27-of-60) |
| Direct Questioning | 50.00% (30-of-60) | 48.33% (29-of-60) | 46.67% (28-of-60) |
| Option Analysis | 45.00% (27-of-60) | 43.33% (26-of-60) | 43.33% (26-of-60) |
| Chain of Thought | 52.54% (31-of-59) | 45.00% (27-of-60) | 41.67% (25-of-60) |
| Answer and Justify | 48.33% (29-of-60) | 50.00% (30-of-60) | 50.00% (30-of-60) |
| Elimination Method | 41.67% (25-of-60) | 46.67% (28-of-60) | 48.33% (29-of-60) |
| Comparative Analysis | 41.67% (25-of-60) | 41.67% (25-of-60) | 43.33% (26-of-60) |
| Contextual Embedding | 48.33% (29-of-60) | 43.33% (26-of-60) | 43.33% (26-of-60) |
| Confidence Scoring | 46.67% (28-of-60) | 53.33% (32-of-60) | 50.00% (30-of-60) |
| Expert Mimicry | 45.00% (27-of-60) | 45.00% (27-of-60) | 45.00% (27-of-60) |
| Consensus Technique | 38.33% (23-of-60) | 43.33% (26-of-60) | 45.00% (27-of-60) |
| Give Model Time to Think | 41.67% (25-of-60) | 46.67% (28-of-60) | 45.00% (27-of-60) |
| Best Prompt (candidate 4) | 46.67% (28-of-60) | 45.00% (27-of-60) | 48.33% (29-of-60) |

**Supplementary Table S5:** The best model temperature was found to be 1, which is a new finding. In LLMs, temperature controls prediction randomness. Lower temperatures yield more predictable text, whereas higher temperatures increase diversity and creativity, potentially at the cost of relevance. [Experiment Setting: 3 runs with max_token= 512plus. Prompt = 'best prompt candidate 4"]

| Model Temperature (temperature) >>E0P3<< | GPT3.5-API with best prompt [a] | GPT3.5-API with raw prompt [a] |
|---|---|---|
| | Average ± SD [95CI] | Average ± SD [95CI] |
| Temperature = 0 | 45.56±0.96 [44.47, 46.65] | 43.89±0.96 [42.8, 44.98] |
| Temperature = 0.1 | 46.11±1.92 [43.93, 48.29] | 45.56±0.96 [44.47, 46.65] |
| Temperature = 0.4 | 45.55±2.55 [42.67, 48.43] | 45.0±1.67 [43.11, 46.89] |
| Temperature = 0.6 | 42.78±3.47 [38.85, 46.71] | 42.78±0.96 [41.69, 43.86] |
| Temperature = 0.9 | 45.0±5.77 [38.46, 51.53] | 45.0±2.89 [41.74, 48.27] |
| Temperature = 1 [selected best temperature] | 47.78±2.55 [44.9, 50.66] | 48.33±2.89 [45.07, 51.6] |

**Footnote:** SD, standard deviation; 95% CI, 95% confidence interval. Raw prompt: "Answer the following question and select one option from the provided options". Best prompt: "Imagine you are a seasoned gastroenterologist approaching this complex question from the gastroenterology board exam. Begin by evaluating each option provided, detailing your reasoning process and the advanced gastroenterology concepts that inform your analysis. The most accurate option is chosen on the basis of your expert judgment, your choice is justified, and your confidence in this decision is rated from 1 to 10, with 10 being the most confident. Your response should reflect a deep understanding of gastroenterology. "

# Supplementary Section 3: Full Details of Experiment One

## 3.1. LLMs in Experiment 1

Experiment 1 aims to show the performance of LLMs on text-based questions. The full details of the LLMs in this experiment are as follows.

**Web interface:**

- GPT-3.5 Web (at the time of use points to the gpt-3.5-turbo-0125 model; March 2024)
- GPT-4 Web (at the time of use points to the gpt-4-0613 model; March 2024)
- Claude3-Haiku Web (fastest Claude3 model claude-3-haiku-20240307; March-April 2024)
- Claude3-Sonnet Web (balanced Claude3 model claude-3-sonnet-20240229; March-April 2024)
- Claude3-Opus Web (most intelligent Claude3 model claude-3-opus-20240229; March-April 2024)
- Gemini Web (Gemini 1.0 Pro model; March 2024)
- Gemini Advanced Web (Gemini 1.0 Ultra model; March-April 2024)

**API:**

- GPT-3.5 API (gpt-3.5-turbo-0125 model; March 2024)
- GPT-4 API (gpt-4-0613 model; March 2024)
- GPT-4o (gpt-4o-2024-05-13; May 2024)
- GPT-4omini (gpt-4o-mini-2024-07-18; July 2024)
- Claude3-Haiku API (claude-3-haiku-20240307; April 2024)
- Claude3-Sonnet API (claude-3-sonnet-20240229; April 2024)
- Claude3-Opus API (claude-3-opus-20240229; April 2024)
- Claude3.5-Sonnet (claude-3-5-sonnet-20240620; July 2024)

**Poe Web Interface - Fully Open-Source Models:**

- Llama2-70b Poe (Llama2-70b-chat; April 2024)
- Llama-2-13b (Llama2-13b-chat; April 2024)
- Llama2-7b Poe (Llama2-7b-chat; April 2024)
- Mistral-Large Poe (mistral-large-2402: this model is not fully open-source; April 2024)
- Mistral-7b (Mistral-7B-Instruct-v0.2; April 2024)
- Mixtral-8x7b Poe (Mixtral-8x7B-v0.1; April 2024)

- Llama3-8b Poe (Llama3-8B-instruct; May 2024)
- Llama3-70b Poe (Llama3-70B-instruct; May 2024)
- Llama3.1-8b Poe (Llama3-70B-instruct; Aug 2024)
- Llama3.1-70b Poe (Llama3-70B-instruct; Aug 2024)
- Llama3.1-405b Poe (Llama3-70B-instruct; Aug 2024)
- Gemma-2-9B (Gemma-2-9B; July 2024)
- Gemma-2-27B (Gemma-2-9B; July 2024)

**Local Computer - Fully Open-Source Models**:

- Mistral-7bQ8 Local (8-bit quantized GGUF version of open-mistral-7B; accessed from huggingface/TheBloke in April 2024)
- Llama2-7bQ8 Local (8-bit quantized GGUF version of llama2-7b-chat; accessed from huggingface/TheBloke in April 2024)
- Llama2-13bQ5KM Local (5-bit-K-M quantized GGUF version of llama2-13b-chat; accessed from huggingface/TheBloke in April 2024)
- MedcineLLM-Q8 Local (8-bit quantized version of medicine-LLM, which is a fine-tuned model based on Llama2-7b; accessed from huggingface/TheBloke in April 2024)
- Llama3-8bQ8 Local (8-bit quantized version of Llama3-8b)
- Phi3-mini3B-fp16 Local (Phi3-16fp mini LLM of phi3-mini-4k-instruct; accessed from microsoft/Phi-3-mini-4k-instruct-gguf in May 2024)
- OpenBioLLM-8bQ8 (8-bit quantized version of OpenBioLLM, which is a fine-tuned version of Llama3-8b)
- Phi3-medium14b-Q6 Local (6-bit quantized version of Phi3-medium with 14B parameters; accessed from bartowski/Phi-3-medium-128k-instruct-GGUF in July 2024)
- Gemma2-9b-Q8 (8-bit quantized Gemma-2-9b; accessed from lmstudio-community/gemma-2-9b-it-GGUF in July 2024)

## 3.2. Local Models, the Quantization Concept, and Computer Configuration

We created a local server via LM Studio and sent requests from Python to the local server to generate responses. LM Studio uses the lllama.cpp python library to empower the user interface for the use of local LLMs. The sources of the local models are described in Supplementary Section 3.1. LLMs used in Experiment 1.

Quantization is a process of making large models that require resources beyond the usual computer, so we can use them on our accessible computers at the cost of lower precision. For example, even a small model with 7 billion (7B) parameters needs approximately 28 GB of memory (7 times 4) for execution, in addition to the memory the system uses for its own. Consequently, not only do big models such as GPT 3.5 (13B), Mixtral8x7 (13.5B), or Llama2-70B, even small models usually with 7B parameters, need cloud computing or supercomputers. Quantization compresses the model size from 32-bit precision to the desired N-bit precision (16). We used an 8-bit quantized version for 7 billion models (Mistral-7b-Q8, Llama2-7b-Q8, and Medcine-LLM-Q8), which requires approximately 8 GB of RAM, and a Q5_K_M version for 13 billion models (Mistral-13b-Q5KM), which requires approximately 12 GB of RAM.

To generate local LLM responses, we used a laptop (Lenovo7 16IAX7) with a 12th Gen Intel Core i9-12900HX with 16 processing cores, 32 GB of DDR5 RAM, and an NVIDIA RTX 3080 Ti laptop GPU with 16 GB of graphic memory.

# Supplementary Section 4: Full Details of Experiment Two

## 4.1 Handling Image-Inclusive Questions

Historical limitations on the availability of image-capable LLMs led previous studies to exclude questions containing images, focusing instead on assessing the models' text-based reasoning capabilities (12,15,16). Our dataset, however, contains a considerable number of questions accompanied by images. To address this, we tested three approaches to image inclusion:

- **Answer without using images**: We evaluated the necessity of images correctly answering their corresponding questions without providing the actual images.
- Prompt: "Imagine you are a seasoned gastroenterologist approaching this complex question from the gastroenterology board exam. Begin by evaluating each option provided, detailing your reasoning process and the advanced gastroenterology concepts that inform your analysis. The most accurate option is chosen on the basis of your expert judgment, your choice is justified, and your confidence in this decision is rated from 1 to 10, with 10 being the most confident. Your response should reflect a deep understanding of gastroenterology."
- **Human-crafted hints:** These descriptions were carefully crafted by two independent gastroenterologists, aiming to provide comprehensive visual information without using high-level classifications or specific terminologies that could inadvertently reveal the answer. For example, an image depicting a perianal abscess was described as "a small perianal bulge with overlying redness and surrounding erythema." Any discrepancies in evaluations or descriptions were resolved through consensus.
- **Prompt**: "Imagine you are a seasoned gastroenterologist approaching this complex question from the gastroenterology board exam. Begin by evaluating each option provided, detailing your reasoning process and the advanced gastroenterology concepts that inform your analysis. The most accurate option is chosen on the basis of your expert judgment, your choice is justified, and your confidence in this decision is rated from 1 to 10, with 10 being the most confident. Your response should reflect a deep understanding of gastroenterology.

You lack direct access to the image. Instead, a human has provided relevant information about images in context. You may use this hint if helpful; otherwise, base your answer solely on the available information."

- **LLM-crafted captions:** The images were provided to LLMs, which are capable of handling images via the prompt "You are an expert doctor with deep knowledge of medical gastrointestinal imaging. Analyze the provided medical image and generate a concise, yet detailed, textual description that captures all essential medical information. This description should enable a multimodal large language model to fully understand the content and context of the image without viewing it. Be helpful, trustworthy, and reliable. Provide a concise image description that can be utilized for diagnosis." Each caption text generated by an LLM family was used by the same LLM member, along with the question and options text.

- **Prompt**: "Imagine you are a seasoned gastroenterologist approaching this complex question from the gastroenterology board exam. Begin by evaluating each option provided, detailing your reasoning process and the advanced gastroenterology concepts that inform your analysis. The most accurate option is chosen on the basis of your expert judgment, your choice is justified, and your confidence in this decision is rated from 1 to 10, with 10 being the most confident. Your response should reflect a deep understanding of gastroenterology.

  You lack direct access to the image. Instead, another LLM has provided relevant information about it. You may use these captions if helpful; otherwise, base your answer solely on the available information."

- **Error handling**: In some cases, the VLMs avoid providing captions. We tagged captions as failed when the LLM avoided providing the caption after three tries. The GPT4V Web, GPT4V API, Claude3OpusV-Web Claude3-OpusV API, and GeminiAdvancedV Web had 0, 19, 2, 3, and 88 errors, respectively.

- **Directly feeding the image:**

- **Prompt:** "Imagine you are a seasoned gastroenterologist approaching this complex question from the gastroenterology board exam. Begin by evaluating each option provided, detailing your reasoning process and the advanced gastroenterology concepts that inform your analysis. The most accurate option is chosen on the basis of your expert judgment, your choice is justified, and your confidence in this decision is rated from 1 to 10, with 10 being the most confident. Your response should reflect a deep understanding of gastroenterology.

Images have been provided for your reference. You may use the information in the images if helpful; otherwise, base your answer solely on the available information."

### 4.2 Models in Experiment 2

The following models were used in experiment 2:

- **GPT4V Web** (Currently points to gpt-4-1106-vision-preview using training data up to Apr 2023; used in April 2024)
- **GPT4V API** (Currently points to gpt-4-1106-vision-preview; used in April 2024)
- **GeminiAdvancedV Web** (Gemini 1.0; used in April 2024)
- **Claude3-OpusV Web** (claude-3-opus-20240229; used in April 2024)
- **Claude3-OpusV API** (claude-3-opus-20240229; used in May 2024)

# Supplementary Section 5: Full Details of the Evaluation

.

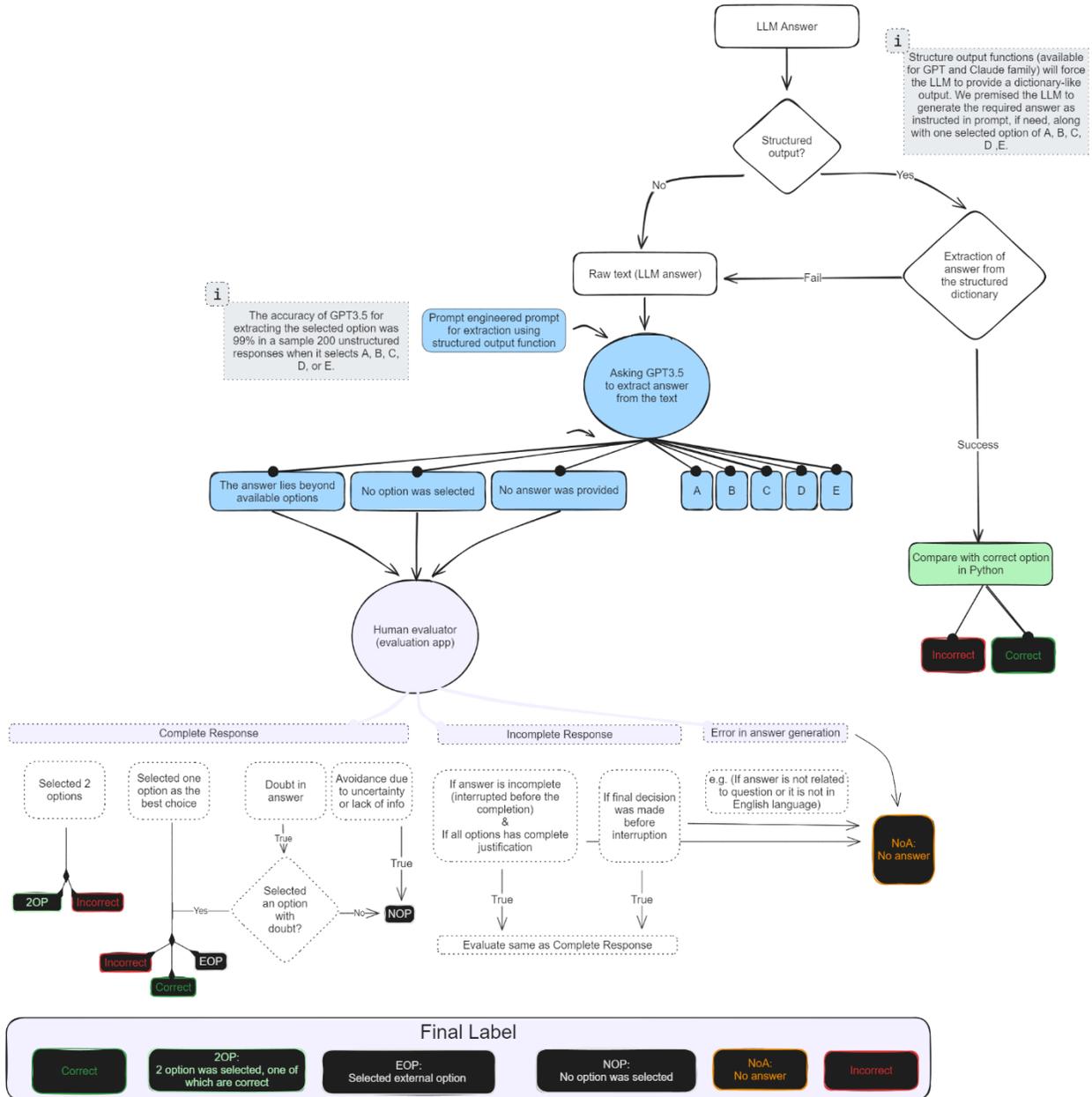

**Supplementary Figure S1.** Details of semiautomated answer extraction from LLMs, including GPT-3.5-based extraction and human evaluation

## 5.1 Extracting Answers from Raw Output Utilizing GPT3.5

For models providing structured output, the correctness of the answer was evaluated by comparing the ground truth and LLM answer via python code. For models without a structured output, the selected option was extracted via the OpenAI extraction schema. The GPT3.5-API has a temperature of 0 and a maximum number of tokens for generation of 50, with the following command:

> **Schema Description**: "Act as a helpful, reliable, and accurate assistant. This response is generated in the context of a multiple-choice question. The LLM was asked to think about each option and then select an option. Your task is to identify the exact chosen option on the basis of the provided text."
>
> **Parameter Description**: "The option selected by the LLM."
>
> **Schema Options**: "A", "B", "C", "D", "E", "No option was selected", "More than one option was selected", "I am not sure."

We tested the performance of this pipeline on a random sample of 200 questions from 4 different answered datasets (Claude3 Haiku-Web, GPT3.5-Web, llama2-7b-Q8-Local, and mixtral-8x7b-Poe). When one of options A to E was provided as the selected option, the accuracy was 99%. The four options of "No option was selected", "More than one option was selected", and "I am not sure." was provided to a human evaluator to label the questions.

## 5.2 Performance of GPT3.5 in Extracting Answers from Raw Textual LLM Responses

To evaluate the performance of GPT-3.5 in extracting selected options from raw responses, we created a dataset comprising 100 answers from various environments. Supplementary Figure S**2** illustrates the performance metrics. GPT-3.5 successfully extracted the correct option in 83% of the cases, achieving an accuracy of 98.8% for those extractions. In the remaining 17% of the cases, GPT-3.5 flagged the questions for manual evaluation, resulting in an overall semiautomated accuracy of 99%.

In Experiment 1, we conducted 33 **LLM** runs on 300 questions from ACG 2022. Eight runs were managed with structured outputs (four GPT runs and four Claude3 runs). Among these structured outputs, 215 out of 2400 responses failed to parse correctly, necessitating manual evaluation. Additionally, across 25 standard outputs using Poe, web interfaces, and local LM studio servers,

GPT-3.5 extracted 8700 responses, 236 of which require human validation. In total, our semiautomated evaluation approach required manual evaluation for 451 out of 11,100 generated answers.

**Supplementary Figure S2. Confusion matrix of GPT-3.5 performance in the extraction of selected options from the raw textual response of LLMs.**

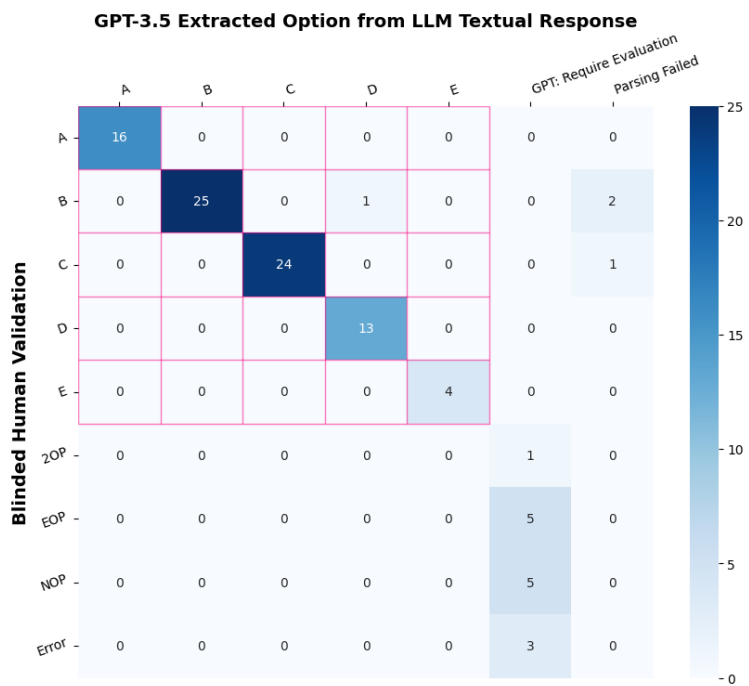

## 5.3 Human Evaluator

One evaluator with a medical doctorate degree and two years of medical practice evaluated answers requiring human evaluation, as mentioned in Supplementary Section 5.2. The guidelines for labeling answers are depicted in **Figure 2**. If one option was selected by LLM, the option was replaced with the LLM answer, and the correctness was evaluated via Python code. If more than one option was selected and the correct answer lies in one of the selected options, the two-option label ('2OP') was added. If no option was selected or the LLM avoided providing any option, no option ('NOP') label was added. If the LLM sought to provide an answer beyond the provided options, the label of the external option ('EOP') was used. In cases where the LLM selected an option while providing some caution or uncertainty, the selected answer was selected as the LLM choice.

## 5.4 Questions Classification: Difficulty, Taxonomy, Care Phase, Subject, Length

**Question Type: Gastroenterology Subject Category**

The 2022 ACG-SA was our main databank for generating answers. The dataset includes 49 questions on the colon, 26 on endoscopy, 36 on the esophagus, 30 on inflammatory bowel disease (IBD), 14 on irritable bowel syndrome (IBS), 52 on the liver, 19 categorized as miscellaneous, 3 on nutrition, 32 on pancreaticobiliary, 22 on the small bowel, and 17 on the stomach.

**Question Type: Cognitive Taxonomy (Bloomberg Taxonomy)**

In this study, we employed a modified Bloom's taxonomy to categorize MCQs on the basis of their required cognitive processes introduced by Nikki Ziadi [44]. This method classifies questions into two levels: lower-order thinking skills (Level 1) and higher-order thinking skills (Level 2). The level 1 questions necessitate the recall or comprehension of learned information without applying a concept, exemplified by questions such as "Which of the following drugs can cause pain, vomiting, and jaundice?" The level 2 questions demand the application of prior knowledge to new information, involving analysis, synthesis, and evaluation. An example is "A 12-year-old girl with sickle cell disease has pain in her right arm. X-ray revealed bony lesions consistent with osteomyelitis. Which of the following is the most likely causal organism?" This categorization allows for a nuanced assessment of LLM capabilities across different cognitive domains.

One author (SAASN) labeled questions via the modified Bloom's taxonomy guide [44]. In the 2022 ACG-SA, only 2 questions targeted lower-order thinking skills. This is expected, as the self-assessment is designed for the gastroenterology board exam, a complex and high-level evaluation. Additionally, 297 questions were case-based, presenting clinical or laboratory scenarios, and 92 questions were integrated, combining multiple domains and disciplines into single test items.

**Question Type: Patient Care Phase**

One author (SAASN) labeled the questions on the basis of their aim to test one or more steps in the patient care process. The questions aimed at evaluating the patient care phase were categorized into five key areas of gastroenterology: 123 questions focused on diagnosis, 217 focused on treatment, 211 focused on investigation, 55 focused on complications, and 3 focused on pathophysiology.

**Question Type: Difficulty using Test-Taker Scores**

The question difficulty was defined as the percentage of respondents who answered the question correctly. The average scores for the human test takers were kindly provided by the ACG upon request. To evaluate performance on the basis of question difficulty, we stratified the questions into 4 quartiles. The questions were classified into Q1 (12.75%-64.92%), Q2 (64.93%-79.23%), Q3 (79.23%-89.44%), and Q4 (89.45%-99.21%), each consisting of 75 questions.

**Question Type: Question Length**

The sum of the question and option tokens was calculated via the tiktoken python library. The token is the smallest unit of data processed by LLMs after translation into the corresponding mathematical representation during the process of embedding. The questions were categorized into three tertiles: short (49--179), medium (180--262), and long (263--588), which resulted in 99, 99, and 102 labels, respectively.

# Supplementary Section 6: Complementary Results

**Supplementary Figure S3.** Granular performance of LLMs (Experiment 1) and question characteristics.

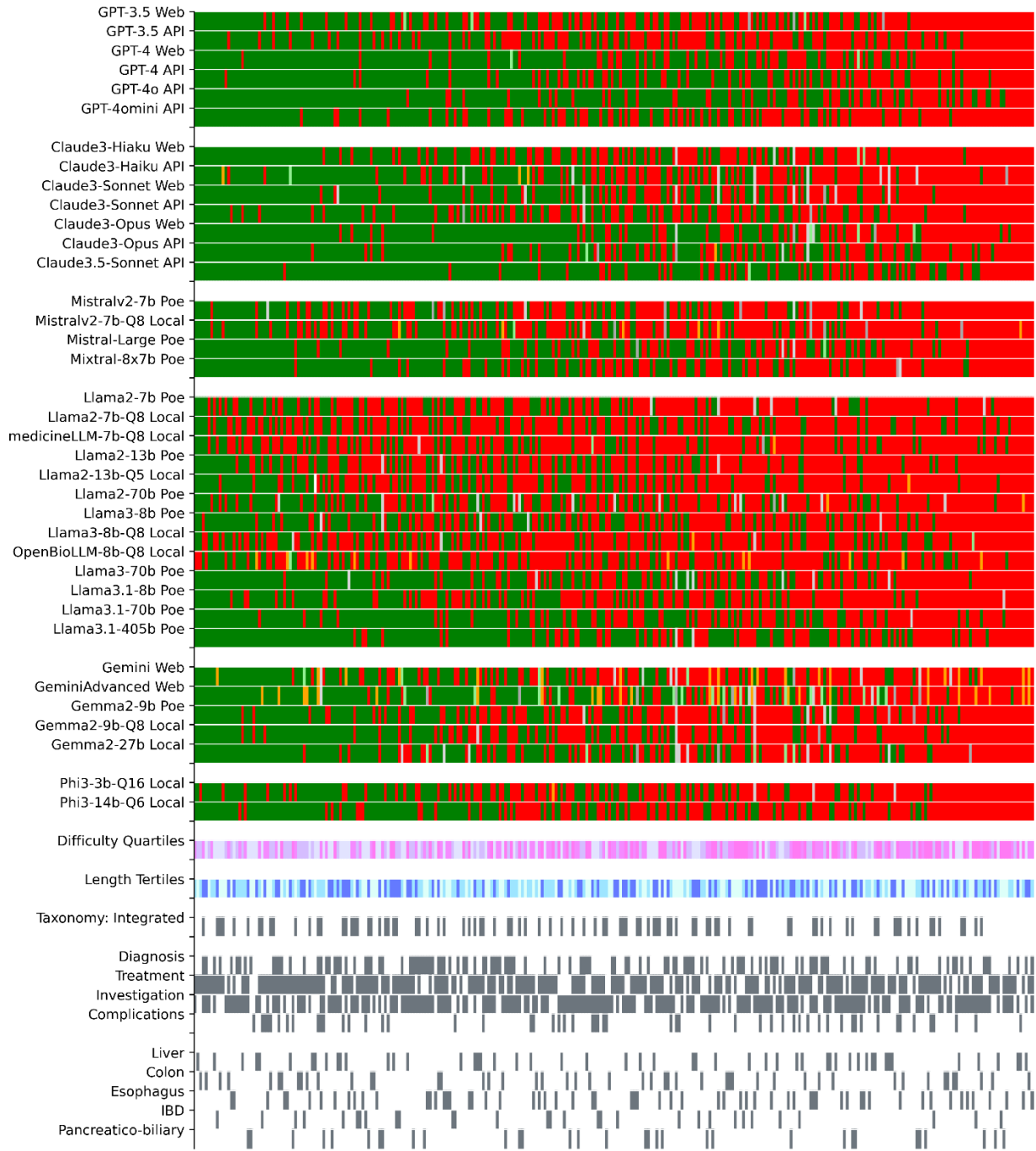

**Supplementary Figure S4. Performance Stratified by Image-Inclusive and Text-Based Questions**

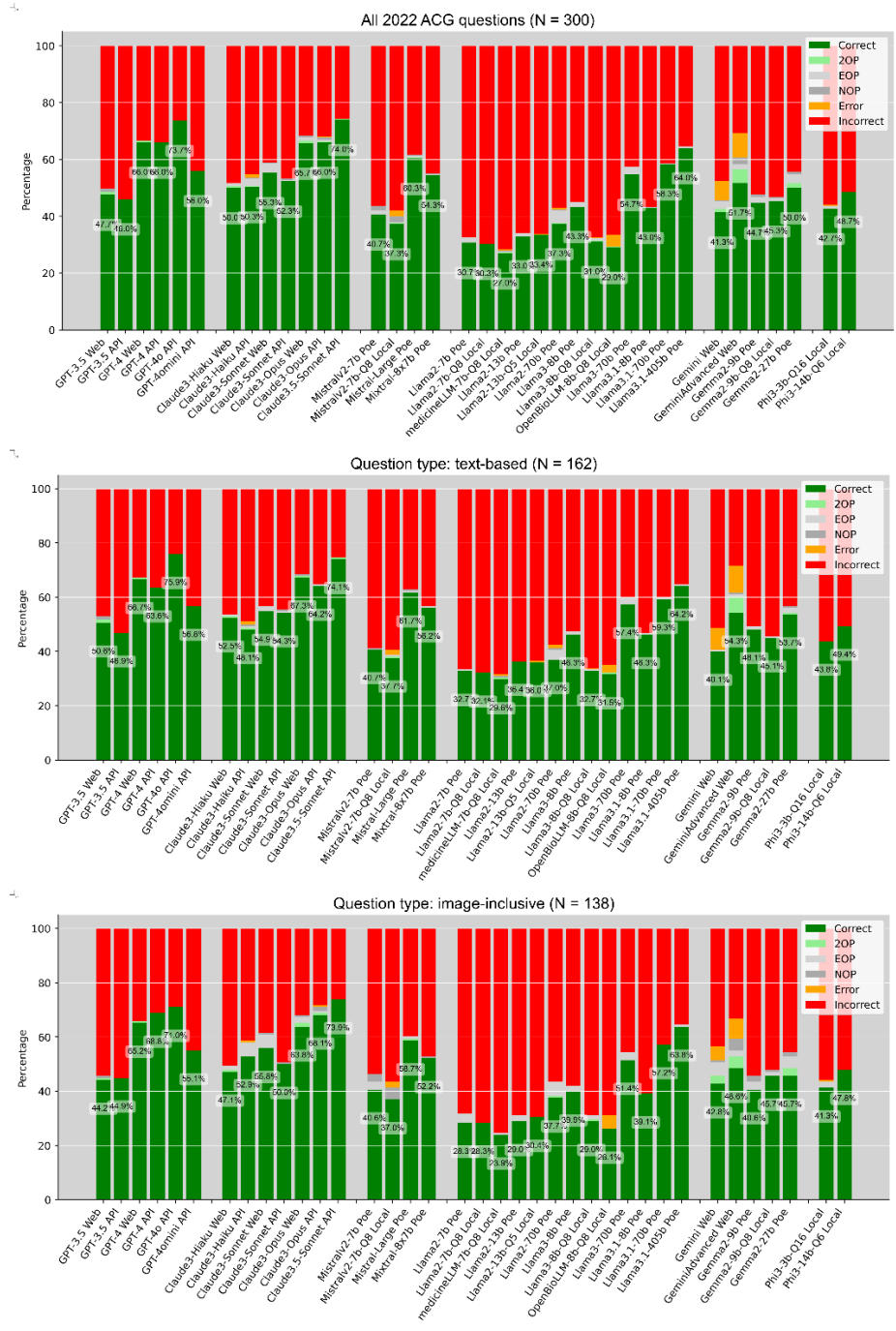

# Supplementary Figure S5. Performance Stratified by Question Subject

# Supplementary Figure S6. Performance stratified by patient care phase

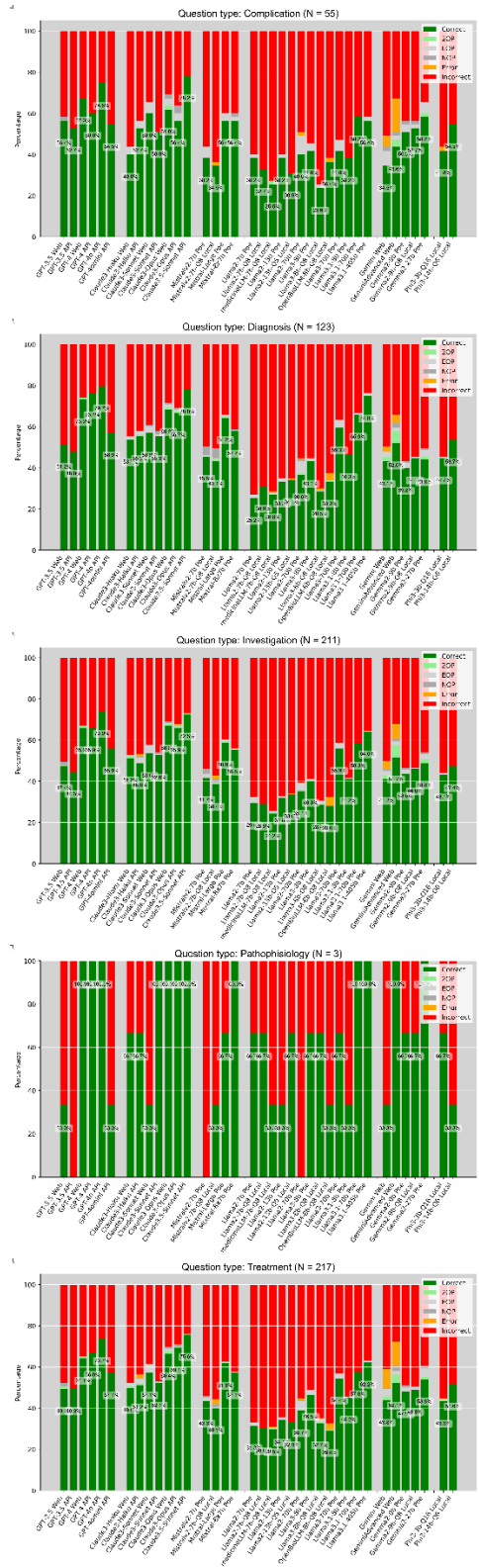

# Supplementary Figure S7. Performance Stratified by Question Difficulty

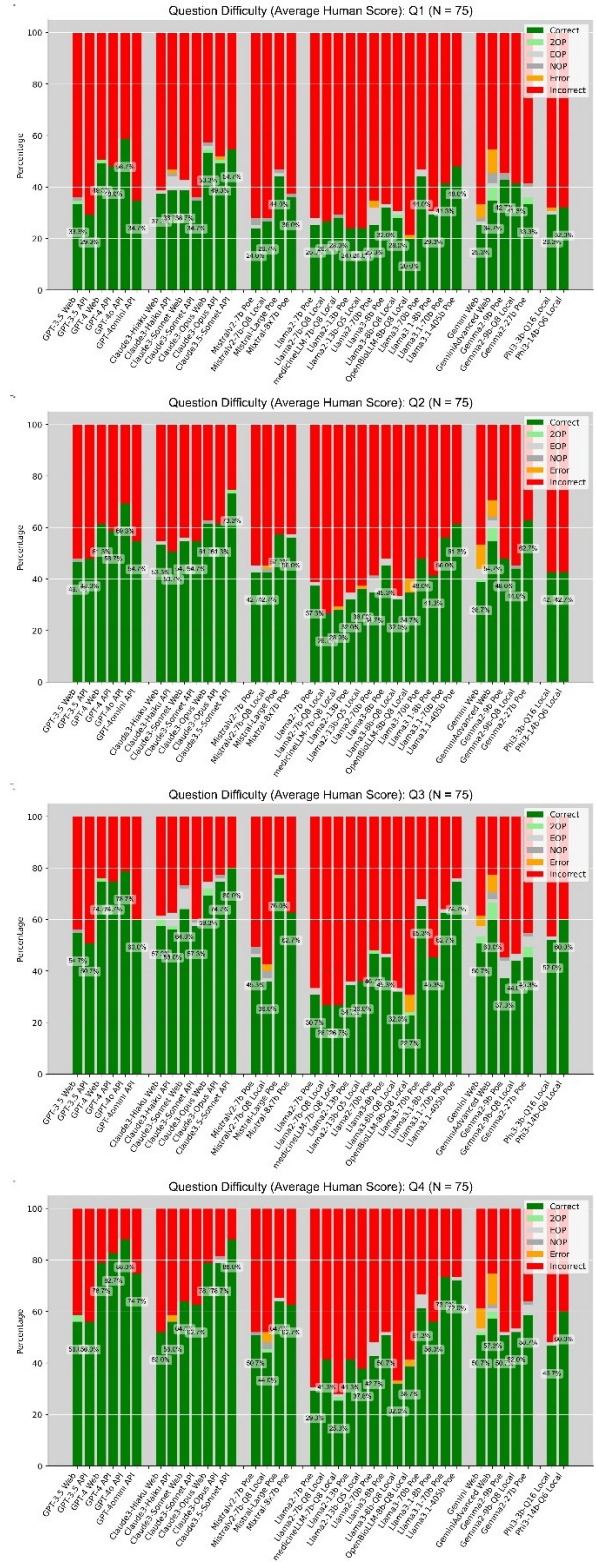

# Supplementary Figure S8. Performance stratified by the question length

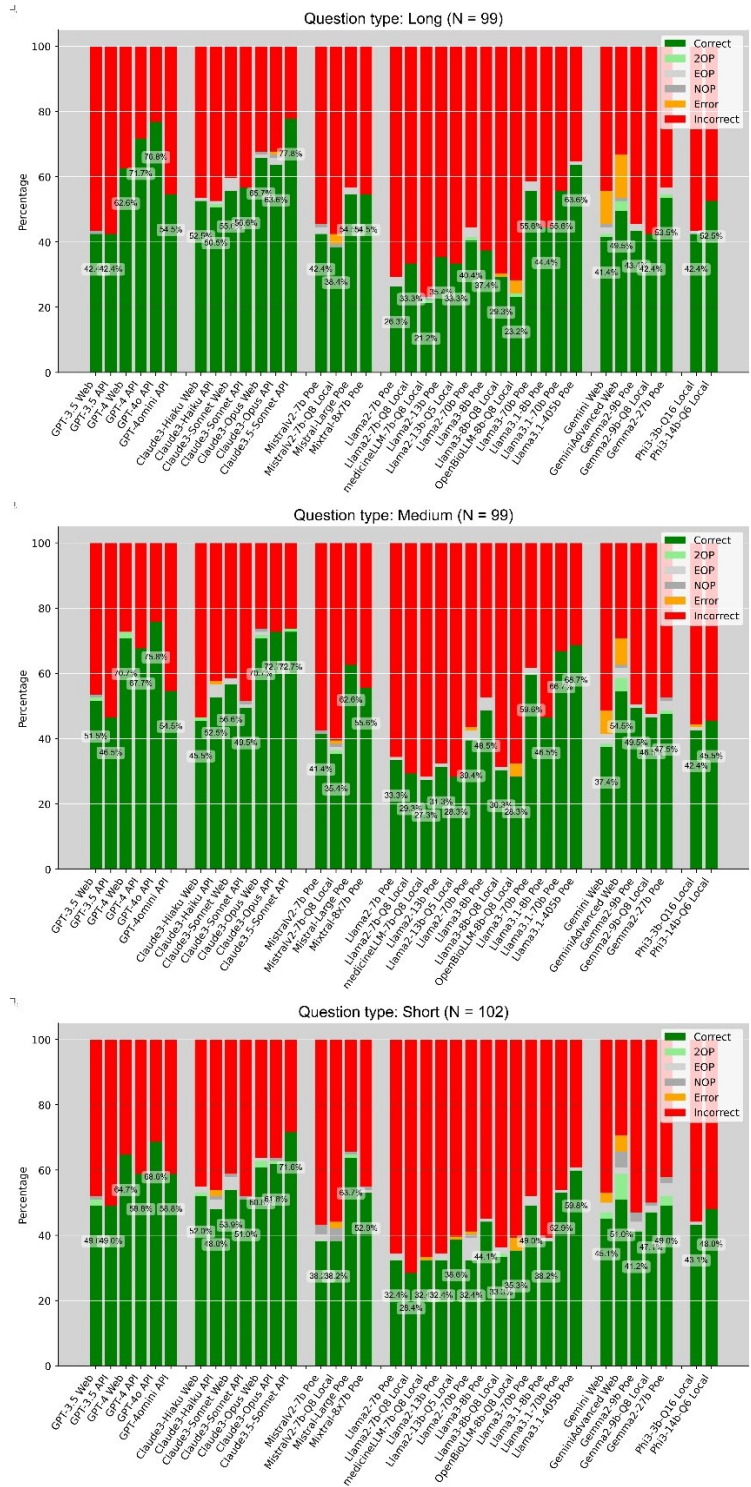

## Supplementary Table S5. The cost and time of executing one question with an average length

| Setup (Cost of Setup) | Model Name | Prompt Token**** | Completion Tokens | Execution Time | Execution Time per 100 Completion Token | Total Cost of 100 similar Execution | Execution Cost per 1 M Prompt Token | Execution Cost per 1 M Completion Token |
|---|---|---|---|---|---|---|---|---|
| **First-Party Web Interface (20$ Free or paid account *)** | | | | | | | | |
| | GPT-3.5 Web | - | 542 | 9 s | 1.66 s | | | |
| | GPT-4 Web | - | 683 | 26 s | 3.81 s□ | | | |
| | GPT-°Web | - | 704 | 12 s | 1.70 s | | | |
| | Claude3-Haiku Web | - | 835 | 8 s | 0.96 s | | | |
| | Claude3-Sonnet Web | - | 681 | 16 s | 2.35 s | | | |
| | Claude3-Opus Web | - | 572 | 31 s□ | 5.42 s | | | |
| | Gemini Web | - | 429 | 10 s | 2.33 s | | | |
| | Gemini Advanced Web | | | | | | | |
| **First-Party API (None)** | | | | | | | | |
| | GPT-3.5 API (Function call) | 426 | 80 | 2 s | 2.50 s | $0.033 | $0.50 | $1.50 |
| | GPT-3.5 Open (Open call) | 426 | 391 | 9 s | 2.30 s | $0.080 | $0.50 | $1.50 |
| | GPT-4 API (Function call) | 423 | 416 | 34 s | 8.17 s | $3.765 | $30.00 | $60.00 |
| | GPT-4 Open (Open call) | 319 | 452 | 31 s□ | 6.86 s | $3.669 | $30.00 | $60.00 |
| | GPT-°API (Function call) | 421 | 489 | 11 s□ | 2.25 s | $0.944 | $5.00 | $15.00 |
| | GPT-°API (Open call) | 317 | 683 | 21 s□ | 3.07 s | $1.183 | $5.00 | $15.00 |
| | Claude3-Haiku API | 649 | 655 | 6 s | 0.92 s | $0.98 | $0.25 | $1.25 |
| | Claude3-Sonnet API | 649 | 548 | 13 s | 2.37 s | $1.017 | $3.00 | $15.00 |
| | Claude3-Opus API | 649 | 597 | 38 s | 6.35 s | $5.451 | $15.00 | $75.00 |
| **Poe Third-Party Web Interface (**Poe Account** 25$) | | | | | | | | |
| | Llama2-70b | - | 669 | 7 s | 1.05 s | | | |
| | Llama2-7b | - | 808 | 2 s | 0.25 s | | | |
| | Mistral-Large | - | 510 | 13 s | 2.55 s | | | |
| | Mixtral-8x7b | - | 507 | 6 s | 1.18 s | | | |
| | Llama3-8b | - | 803 | 9 s | 1.12 s | | | |
| | Llama3-70b | - | 612 | 14 s | 2.29 s | | | |
| **Local Machine**** (~2500$ Laptop or 1900$ Desktop Machine) | | | | | | | | |
| | Mistral-7bQ8 | 385 | 511 | 47 | 9.20 s | | | |
| | Llama2-7bQ8 | 414 | 436 | 39 | 8.94 s | | | |
| | Llama2-13bQ5KM | 412 | 483 | 96 | 19.88 s | | | |
| | MedcineLLM-Q8 (Llama2-7b fine-tuned version) | 465 | 511 | 47 | 9.20 s | | | |
| | Llama3-8bQ8 Local | 332 | 221 | 44 | 19.91 s | | | |
| | OpenBioLLM-Q8 (Llama3-8b fine-tuned version) | 339 | 154 | 22 | 14.29 s | | | |
| | Phi3-fp16 Local | 402 | 511 | 47 | 9.20 s | | | |

*Footnote: The same model settings as those used in Experiment 1 were used. The execution time was recorded for one sampled question, which was defined as the time from when the participant pressed to the time of the last token. We used tiktoken to count the tokens for all model outputs. Token prices were extracted on June 20, 2024, from [https://openai.com/api/pricing/](https://openai.com/api/pricing/) and [https://www.anthropic.com/api](https://www.anthropic.com/api)*

*\*: The price of a paid account for OpenAI, Claude3, and Gemini services was $20/month. GPT-4 and Gemini Advanced are accessible only via paid subscriptions, whereas other models can be run on a free account but with considerable request limits. \*\*: Poe provides 1,000,000 computation tokens for one month with a subscription. Sending each prompt incurs a specified number of computation tokens. \*\*\*: The local machine specifications were a GPU RTX 3080 (16 GB graphic memory) and a CPU i9--12900HX (32 GB DDR5 4800 RAM). We ran the local models using 16 layers offloaded to the GPU and 10 CPU threads. \*\*\*\*: In both the API and local environments, the model returns the number of input tokens (i.e., prompt tokens). When function calls are used, the schema is injected into the prompt, resulting in a greater number of prompt tokens.*